\documentclass[10pt, a4paper]{article}

\usepackage[]{lrec2026} 
\usepackage{booktabs}
\usepackage{makecell}
\usepackage{tabularx}
\usepackage{todonotes}
\usepackage{subfig}
\usepackage{graphicx}

\usepackage{amsmath}
\usepackage{multirow}

\title{Merge and Conquer: Instructing Multilingual Models by \emph{Adding} Target Language Weights}

\name{Eneko Valero, Maria Ribalta i Albado, Oscar Sainz, Naiara Perez, German Rigau} 

\address{\\HiTZ Center - Ixa, University of the Basque Country UPV/EHU\\
         \{name.surname\}@ehu.eus\\}

\abstract{
Large Language Models (LLMs) remain heavily centered on English, with limited performance in low-resource languages. Existing adaptation approaches, such as continual pre-training, demand significant computational resources. In the case of instructed models, high-quality instruction data is also required, both of which are often inaccessible for low-resource language communities. Under these constraints, model merging offers a lightweight alternative, but its potential in low-resource contexts has not been systematically explored.
In this work, we explore whether it is possible to transfer language knowledge to an instruction-tuned LLM by merging it with a language-specific base model, thereby eliminating the need of language-specific instructions and repeated fine-tuning processes whenever stronger instructed variants become available. Through experiments covering four Iberian languages (Basque, Catalan, Galician, and Spanish) and two model families, we show that merging enables effective instruction-following behavior in new languages and even supports multilingual capability through the combination of multiple language-specific models. Our results indicate that model merging is a viable and efficient alternative to traditional adaptation methods for low-resource languages, achieving competitive performance while greatly reducing computational cost.
 \\ \newline  }

\begin{document}

\maketitleabstract

\section{Introduction}

Recent advances in Large Language Models (LLMs) have significantly improved their multilingual capabilities. Modern LLMs, particularly those deployed as commercial products, are generally expected to understand and generate text in high-resource languages such as English, Chinese, and Spanish. However, this performance does not extend uniformly across all languages. For low-resource  languages, especially those with limited online presence, LLMs continue to exhibit substantial performance degradation, even in state-of-the-art models~\citep{moroni-etal-2025-multi,baucells-etal-2025-iberobench,grandury-etal-2025-la}. This disparity can be attributed to both the scarcity of training data and the limited economic incentives to support these languages, resulting in inconsistent and often unreliable behavior from frontier models. In an attempt to improve the LLM capabilities for certain languages, several works have presented different approaches, mainly based on continual pre-training, to adapt already trained LLMs to low-resource languages~\cite{etxaniz-etal-2024-latxa,ustun2024ayamodelinstructionfinetuned}.

The emergence of instruction-tuned and aligned LLMs has further raised the bar for low-resource model development. Beyond requiring large-scale corpora in the target language, these models also depend on high-quality instruction--response pairs to guide behavior during fine-tuning. For low-resource languages, already constrained by limited textual data, curating such instruction datasets is particularly challenging, and, in some cases, practically infeasible. Fortunately, recent work demonstrates that combining target-language corpora with English-language instructions can yield competitive models for low-resource languages~\cite{sainz2025instructinglargelanguagemodels}. However, this approach introduces a new bottleneck: each time a more capable instructed model is released, the target language community must repeat the adaptation process through continued pretraining or fine-tuning. As with data availability, computational resources are often scarce in these communities, further limiting their ability to keep pace with rapid advancements in LLM development.

Motivated by the goal of enhancing LLM capabilities in low-resource languages while simultaneously reducing the substantial computational cost typically required, this paper explores the use of model merging techniques as a more compute efficient alternative to joint pre-training~\cite{sainz2025instructinglargelanguagemodels}. Specifically, we investigate whether it is possible to teach a new language to an already instructed LLM through weight merging. That is, we aim for a model as proficient in the target language as the language-specific base model, while maintaining the instruction following capabilities of the instructed variant. This approach allows adapting any newly released and potentially stronger instructed variant of a base LLM to a target language, while requiring the base model to be trained on that language only \emph{once}. Our experiments, conducted on four languages with varying resource levels (namely, Basque, Galician, Catalan, and Spanish) and across two model families, demonstrate that this is indeed feasible. Moreover, we show that a single LLM can acquire multiple languages by merging several language-specific models. In addition, to evaluate the instruction-following competence of the merged models, we extended the IFEval~\cite{zhou2023instructionfollowingevaluationlargelanguage} benchmark currently available to Spanish and Catalan, to Basque and Galician languages.

In sum, this paper makes the following contributions. First, we demonstrate that it is possible to develop language-specific instructed LLMs by merging language-specific base models with general instructed models. Second, we conduct an in-depth analysis of the behavior of various merging techniques and hyperparameter configurations within our setup. Finally, we release both base and instructed models for Basque, Galician, Catalan, and Spanish developed during our experiments; and, the Basque version of the IFEval dataset.\footnote{\url{https://hf.co/collections/HiTZ/merge-and-conquer}}

\section{Related Work}

Developing LLMs for under-resourced languages remains a major challenge due to the extensive data and computational requirements that are needed to pre-train and post-train models from scratch. As a result, most research has moved towards multilingual pretraining, where a single model is exposed to data from many languages~\cite{le-scao-etal-2022-language,shliazhko2023mgptfewshotlearnersmultilingual}. While this strategy provides broad coverage, the percentage of data destined to low-resource languages is usually insignificant, and often results in poor performance~\cite{Gonzalez2025IberBenchLEB,Bao2023ConversationsIGC}. This imbalance is clear in large-scale web corpora: according to Common Crawl statistics,\footnote{\url{https://commoncrawl.github.io/cc-crawl-statistics/plots/languages}} English alone accounts for nearly half of the available text, whereas Basque represents barely 0.03\%, placing it under the low-resource category. This uneven distribution explains the limited performance of multilingual LLMs on Iberian languages despite their broad coverage. As observed in recent work, Spanish enjoys abundant resources, while Basque, Galician, and Catalan remain largely under-represented, limiting their ability to benefit from the capabilities observed in state-of-the-art LLMs~\cite{etxaniz-etal-2024-latxa,de-dios-flores-etal-2024-corpusnos}. Consequently, multilingual pretraining alone proves insufficient when it comes to low-resource languages.  

Multilingual instruction tuning has become a main strategy for instructing models to low-resource languages, complementing the pretraining stage with language-specific instructions. To address the scarcity of native instruction data, researchers have proposed several approaches such as translating existing datasets, generating synthetic examples, or leveraging English-centric corpora for cross-lingual transfer. For instance, Aya~\citep{ustun2024ayamodelinstructionfinetuned} extends instruction tuning to over 100 languages, more than half of which are low-resource, through large multilingual finetuning. Other works improve data quality via translation-based methods like \cite{nguyen2024betteralignmentinstructionbackandforth}, which builds high-quality instruction-response pairs through iterative translation and rewriting. In the Iberian context, \citet{Bao2023ConversationsIGC} present a curated Galician model, and the Salamandra initiative \cite{gonzalezagirre2025salamandratechnicalreport} integrates resources for Catalan, Galician, Spanish, and Basque. Alternatively, \citet{sainz2025instructinglargelanguagemodels} explored systematically all possible approaches to instruct or adapt a LLM for a specific low-resource language. They proved that joint instruction tuning, even with just English instructions, and continual pre-training substantially outperforms traditional approaches. However, while these efforts show a promising performance, they remain computationally expensive because they must be repeated with every new model release. 


Model merging has recently emerged as an alternative to traditional multi-task learning, offering an efficient way to combine model capabilities from different experts without additional training~\cite{10.1145/3787849}. These approaches interpolate parameters between two or more models using different methods. \textit{Task Arithmetic} builds task vectors from fine-tuned models and combines them additively~\citep{ilharco2023editingmodelstaskarithmetic}, while DARE drops and rescales delta parameters to reduce interference during merging~\cite{Yu2023LanguageMAA}, and TIES prunes redundant parameters across models~\cite{Yadav2023TIESMergingRIA}. 
Research on low-resource multilingual scenarios is still emerging and remains limited with few works showing promising results~\citep{Tao_2024,huang2024chatvector,pipatanakul2025adaptinglanguagespecificllmsreasoning}.
More recently, \citet{cao2025paramdelta} and \citet{sarasua2025diploma} applied the Task Arithmetic merging approach to transfer instruction-following capabilities to continued pre-trained models, with the latter placing special emphasis on low-resource languages such as Basque. 
Nevertheless, these works do not provide a systematic analysis and evaluation of the different alternatives within the merging paradigm for low-resource languages. In contrast, our study focuses on a broad experimental characterization of language transfer via merging across four Iberian languages, model families, sizes, and merging strategies, and evaluates both language competence (through multiple-choice benchmarks and machine translation) and instruction-following behavior (IFEval).

\section{Methodology}

Adapting an instructed LLM to a target language using merging techniques follows a straightforward methodology. First, a base model must be trained to be proficient in the target language. Then, this newly trained base model is merged with an existing instructed model using a merging technique. This section describes the construction of the components used in our experiments. We begin by detailing the available resources, followed by the procedures for training the base LLMs and performing the model merging.

\subsection{Continued pre-training}

\begin{table}[]
    \centering
    \resizebox{\linewidth}{!}{
        \begin{tabular}{l|ccc}
            \toprule
            Language      & Documents & Llama 3.1 & Qwen3 \\ \midrule
            Basque (eu)   & 4.2M      & 3.5B      & 3.5B  \\
            Galician (gl) & 8.9M      & 3.5B      & 3.5B  \\
            Catalan (ca)  & 3.8M      & 3.7B      & 3.8B  \\
            Spanish (es)  & 3.8M      & 3.4B      & 3.5B  \\ \midrule
            English (en)  & 0.5M      & 0.3B      & 0.3B  \\
            \bottomrule
        \end{tabular}
    }

    \caption{Corpus sizes for each language in documents, Llama 3.1 tokens and Qwen3 tokens.}
    \label{tab:corpus-size}
\end{table}

We trained language-adapted base models for four Iberian languages: Basque, Galician, Catalan and Spanish. To train language-specific base LLMs, we followed the methodology proposed by ~\citet{etxaniz-etal-2024-latxa}, who used a Basque corpus comprising approximately 4.2 billion tokens. To enable fair comparisons across languages, we limited the corpus size for all the languages to roughly the same number of tokens. Table~\ref{tab:corpus-size} presents the corpus statistics in terms of document counts, as well as token counts for Llama 3.1 and Qwen 3 (corresponding to the models used in our experiments; \S~\ref{sec:experimental-setup}). As expected, token counts vary slightly depending on the tokenizer, but remain comparable in overall size. Note that we included a small-sized English corpus, which was first proven essential to avoid catastrophic forgetting in~\citet{etxaniz-etal-2024-latxa}, and later confirmed in ~\citet{elhady-etal-2025-emergent}. 

\paragraph{Corpus collection.} For Basque, we use the pretraining data from the Latxa corpus~\cite{etxaniz-etal-2024-latxa}, which consists of 4.3M documents and 1.2B words (mainly massive web-crawl content, news pieces, and encyclopedic text). In the case of Galician, we rely on the CorpusNÓS corpus~\cite{de-dios-flores-etal-2024-corpusnos} of 9.7M documents and 2.1B words drawn from web crawls and public administrations, among others. Spanish and Catalan data are taken from the massive, multilingual CulturaX corpus~\cite{nguyen-etal-2024-culturax}. Given the substantially larger size of CulturaX compared to the Basque and Galician resources, we implemented a series of strategies to obtain a more targeted subset. For Spanish, we retained only documents whose URLs contain a top-level domain indicating origin in Spain (namely, \texttt{.es}, \texttt{.eus}, \texttt{.cat}, or \texttt{.gal}). In addition, both Spanish and Catalan data were filtered using the Dolma toolkit~\cite{soldaini-etal-2024-dolma}, with the pre-implemented Gopher~\cite{rae2021scaling} and C4~\cite{raffel2020exloring} heuristics. For the English subset, we sampled 500k documents from the FineWeb corpus~\cite{penedo2024the}.

\paragraph{Model training.} We trained the models with a sequence length of 8,196 tokens and an effective batch size of 256 instances, corresponding to a total of approximately 2 million tokens per optimization step. Training employed a cosine learning rate scheduler with a warm-up ratio of 0.1 and a peak learning rate of $1\times10^{-5}$. Experiments were conducted on the CINECA Leonardo high-performance computing cluster, utilizing 32 nodes, each equipped with 4 NVIDIA A100 GPUs (64 GB memory). For distributed training, we adopted Fully Sharded Data Parallel \citep{zhao2023pytorchfsdpexperiencesscaling}, which shards model parameters, optimizer states, and activations across all GPUs to maximize memory efficiency and scalability.

\subsection{Model merging}

Model merging refers to the process of combining two or more models, typically sharing the same base architecture and initialization, into a single unified model whose parameters $\theta_{\text{merge}}$ are derived from the parameter sets $\{\theta_1, \theta_2, \cdots, \theta_n\}$ of individual \textit{expert} models. Formally, the simplest instance of model merging can be expressed as a weighted linear interpolation of model parameters, i.e.,
\[
\theta_{\text{merge}} = \sum_{i=1}^{n} w_i \, \theta_i,
\]
where the scalar weights $w_i$ sum to one~\citep{wortsman2022modelsoupsaveragingweights}. While such naive interpolation methods can already transfer useful knowledge between models, more sophisticated approaches explicitly address parameter alignment, conflict resolution, or parameter importance weighting to mitigate performance degradation when models have diverged substantially during training~\citep{ilharco2023editingmodelstaskarithmetic,Yu2023LanguageMAA}. These methods aim to approximate the effect of multi-task training without requiring access to the original fine-tuning data, making them particularly valuable for domains or languages where annotated resources are scarce. In this work, we evaluate several merging strategies across two complementary experimental settings: \textbf{monolingual merging} and \textbf{multilingual merging}.

\paragraph{Monolingual Merging.} 
In the monolingual setting, we study how different merging techniques perform when adapting a model to a new language while preserving previously learned capabilities. Specifically, we analyze the impact of various merging algorithms and their associated hyperparameters on the transfer efficiency and stability of the resulting model. This allows us to assess the extent to which model merging can serve as an alternative to conventional fine-tuning when incorporating additional linguistic knowledge.

\paragraph{Multilingual Merging.} 
In the multilingual setting, we investigate how multiple monolingual expert models can be effectively combined to produce a single multilingual, instruction-tuned LLM. We examine how different merge ratios and strategies influence the balance between languages, evaluating whether specific methods favor stronger per-language specialization or instead yield more uniform cross-lingual performance. This analysis provides insights into the ability of merging techniques to fuse diverse linguistic competencies into a unified model without the need for costly joint training.

\section{Experimental Setup}
\label{sec:experimental-setup}

To thoroughly evaluate our hypotheses, we conducted experiments across multiple languages (Basque, Galician, Catalan, and Spanish), diverse model families (\S~\ref{ssec:models-and-baselines}), various merging methods (\S~\ref{ssec:merge-techniques}), and several benchmarks (\S~\ref{ssec:evaluation-benchmarks}). The following sections provide a detailed description of our experimental setup.

\subsection{Models and baselines} \label{ssec:models-and-baselines}

Our experimental design requires that both the base and instructed variants of each model family be publicly available. Consequently, we selected two widely used and representative model families: Llama 3.1~\cite{grattafiori2024llama3herdmodels} and Qwen 3~\cite{yang2025qwen3}. Llama 3.1 is a well-established model family within the community, known for its strong multilingual performance and widespread adoption. In contrast, Qwen 3 exemplifies the recent emergence of model families that are not predominantly centered on English, reflecting a growing shift towards more linguistically diverse large language models. Given our computational constraints and the high cost associated with large-scale training, we opted to focus on moderate-sized language models. Specifically, we used the 8B variants of both Llama 3.1 and Qwen 3, as well as the 14B variant of Qwen 3, striking a balance between performance and feasibility for our experiments.

Regarding the baselines, we considered two main points of comparison. The first are the existing instructed variants of the chosen model families (namely, Llama 3.1 Instruct and Qwen 3 Instruct) which serve as our non–language-adapted but instructed baselines, allowing us to measure how much the instruction following capabilities are retained after merging. Our second baseline is the language-adapted but not instructed baseline, which allows us to test the language proficiency transfer of the methods. Additionally, we compare our approach to the current state-of-the-art method for low-resource language adaptation, which jointly combines continued pre-training and instruction tuning in a single training phase~\cite{sainz2025instructinglargelanguagemodels}. Specifically, we compare our approach against their best model release for Basque.\footnote{\url{https://huggingface.co/HiTZ/Latxa-Llama-3.1-8B-Instruct}} Moreover, we include in the comparison Salamandra and ALIA models~\citep{gonzalezagirre2025salamandratechnicalreport}, two LLMs specifically trained for our target Iberian languages.

\subsection{Merge techniques} \label{ssec:merge-techniques}

Several model merging techniques have been proposed in the literature. In this work, we focus on four representative and conceptually simple approaches: \textit{Linear Merging}, \textit{Task Arithmetic}, \textit{DARE} and \textit{Breadcrumbs}.

\paragraph{Linear Merging.} 
The \textit{Linear} method~\cite{wortsman2022modelsoupsaveragingweights} is the most straightforward approach to model merging. It consists in performing a weighted average of the parameters from multiple models, assigning a specific coefficient to each. This technique can be viewed as an extension of the idea behind model soups, where model weights are interpolated to combine knowledge from different checkpoints or fine-tuned variants.

\begin{equation}
    \begin{split}
        \mathcal{M}_{\text{Linear}}(\{\theta_i\}_{i=1}^N, \theta_{\text{base}}) 
        = w_{\text{base}}\cdot\theta_{\text{base}} 
        + \sum_{i=1}^N w_i\cdot\theta_i, \\
        \text{where} \quad 
        w_{\text{base}} + \sum_{i=1}^N w_i = 1.
    \end{split}
\end{equation}

\paragraph{Task Arithmetic.} 
The \textit{Task Arithmetic} method, introduced by~\citet{ilharco2023editingmodelstaskarithmetic}, is based on the notion of \textit{task vectors}. For a given task $t_i$, the task vector $\tau_i$ is defined as the difference between the task-specific model and the base model, i.e., $\tau_i = \theta_i - \theta_{\text{base}}$. These task vectors capture the parameter updates associated with adapting the base model to a specific task. Model merging is then performed by adding a weighted combination of these vectors to the base model:

\begin{equation}
    \mathcal{M}_{\text{TA}}(\{\theta_i\}_{i=1}^N, \theta_{\text{base}}) 
    = \theta_{\text{base}} + \sum_{i=1}^N w_i\cdot(\theta_i - \theta_{\text{base}}).
\end{equation}

\paragraph{TIES.}
The \textit{Trim and Elect (TIES)} method~\cite{Yadav2023TIESMergingRIA} offers an alternative to the standard computation of the task vector $\tau_i$. It addresses two main challenges: update redundancy and parameter sign disagreement. The first issue is mitigated by setting to $0$ those values in $\tau_i$ that fall below the top $k$\%, through the application of a mask $m^k_i$ to the task vector $\tau_i$. To handle sign conflicts, the method computes an aggregate \textit{elected sign vector} $\gamma_{m}$ by taking the sign of the average values across the different task vectors. Finally, only the parameters consistent with the elected sign are retained for the final task vector $\hat{\tau}_i$. Formally,

\begin{equation}
    \begin{split}
        \tau'_i &= m^k_i \odot \tau_i, \\
        \gamma_{m,p} &= \text{sgn}\!\left(\sum_{t=1}^n \tau'_{i,p}\right), \\
        \mathcal{A}_p &= \{t \in [n] \;|\; \text{sgn}(\tau'_{i,p}) = \gamma_{m,p}\}, \\
        \hat{\tau}_{i,p} &= \frac{1}{|\mathcal{A}_p|}\sum_{t\in\mathcal{A}_p} \tau'_{i,p}
    \end{split}
\end{equation}

\noindent In our experiments, we did not use the TIES method directly, but applied it in conjunction of the DARE and Breadcrumbs methods defined below. 

\paragraph{DARE.}
The \textit{Drop And REscale} (DARE) method~\cite{10.5555/3692070.3694452} aims to sparsify task vectors in order to reduce interference between tasks during model merging. Similar to Task Arithmetics, it operates on the difference between a fine-tuned model and its base model, defined as $\tau_i$. However, it proposes to randomly drop a proportion $p$ of its parameters and rescale the remaining ones by a factor of $\frac{1}{1-p}$ to preserve the expected magnitude. 
Formally, DARE produces a sparsified task vector:
\begin{equation}
    \tau_i^{\text{DARE}} = \frac{(1 - Z_i) \odot \hat{\tau}_i}{1 - p},
\end{equation}
where $Z_i$ is a binary mask sampled element-wise from a Bernoulli distribution with parameter $p$, and $\odot$ denotes element-wise multiplication. 
The merged model is then obtained as:
\begin{equation}
    \mathcal{M}_{\text{DARE}}(\{\theta_i\}_{i=1}^N, \theta_{\text{base}}) 
    = \theta_{\text{base}} + \sum_{i=1}^N w_i \cdot \tau_i^{\text{DARE}}.
\end{equation}

\paragraph{Model Breadcrumbs.}~\citet{10.1007/978-3-031-73226-3_16} provides a deterministic alternative to DARE by filtering both negligible and extreme parameter updates in the task vectors. 
For each layer of a task vector $\tau_i$, values below a lower threshold (small perturbations) and above an upper threshold (outliers) are masked out, retaining only the mid-range updates that are considered most informative. 
Formally, let $f(\tau_i)$ denote this layer-wise filtering operation; the merged model is computed as:
\begin{equation}
    \begin{split}
        \mathcal{M}_{\text{BC}}(\{\theta_i\}_{i=1}^N, \theta_{\text{base}}) 
        = \theta_{\text{base}} + \sum_{i=1}^N w_i \cdot f(\hat{\tau}_i), \\
        f^L(\hat{\tau}_i) =
        \begin{cases}
            \hat{\tau}^L_{i}, & \text{if } \gamma^L \leq |\hat{\tau}_{i}^L| \leq \beta^L, \\
            0, & \text{otherwise.}
        \end{cases}
    \end{split}
\end{equation}

\subsection{Evaluation benchmarks} \label{ssec:evaluation-benchmarks}

We conducted our evaluations using the LM Evaluation Harness framework~\cite{eval-harness}. Each model variant was tested on a suite of benchmarks on five languages: Basque, Galician, Catalan, Spanish, and English. Our evaluation setup includes multiple-choice benchmarks together with machine translation and instruction-following datasets. We incorporated machine translation and instruction-following as they are text generation tasks, allowing us to better assess the models' language generation capabilities. In total, we conducted evaluations in 14 different benchmarks and their language variants, if available. All of the results are obtained by prompting the models with 5 examples (i.e., 5-shot evaluation).

\paragraph{Multiple-choice benchmarks.} Our evaluation framework includes tasks from multiple categories to assess a range of language understanding and generation capabilities. For \textit{reading comprehension}, we used \textbf{Belebele}~\cite{bandarkar-etal-2024-belebele} and \textbf{EusReading}~\cite{etxaniz-etal-2024-latxa}; the former is available in all evaluation languages, while the latter is specific to Basque. To evaluate \textit{common sense reasoning}, we employed \textbf{XStoryCloze}~\cite{lin-etal-2022-shot}, which is also available in all our target languages. \textit{Linguistic proficiency} in Basque was assessed using \textbf{EusProficiency}~\cite{etxaniz-etal-2024-latxa}, while \textit{linguistic acceptability} was evaluated using language-specific variants of \textbf{CoLA}~\cite{warstadt-etal-2019-neural}: \textbf{GalCoLA}~\cite{baucells-etal-2025-iberobench} for Galician, \textbf{CatCoLA}~\cite{PLN6609} for Catalan, and \textbf{EsCoLA}~\cite{bel-etal-2024-escola} for Spanish. For \textit{miscellaneous knowledge}, we used \textbf{BertaQA}~\cite{NEURIPS2024_3bb42f6b}, \textbf{EusTrivia} and \textbf{EusExams}~\cite{etxaniz-etal-2024-latxa} in Basque, and \textbf{OpenBookQA}~\cite{mihaylov-etal-2018-suit} for Galician, Catalan, and Spanish. Lastly, to evaluate \textit{paraphrasing capabilities}, we used \textbf{Paráfrases}~\cite{baucells-etal-2025-iberobench} in Galician. For all the classification tasks we have used accuracy as our evaluation metric.

\paragraph{Machine Translation.} We include machine translation as a text generation task to better assess model performance in low-resource languages, where grammatical correctness is often lacking~\cite{sainz2025instructinglargelanguagemodels}. For this purpose, we use the Flores benchmark~\cite{goyal2021flores101evaluationbenchmarklowresource} and group languages into two categories: high resource (Spanish, English) and low resource (Basque, Galician, Catalan). We evaluate translation in both directions: $\{\text{es}, \text{en}\} \rightarrow \{\text{eu}, \text{gl}, \text{ca}\}$ and $\{\text{eu}, \text{gl}, \text{ca}, \text{es}, \text{en}\} \rightarrow \{\text{es}, \text{en}\}$. All results are reported using BLEU scores, averaged across the target languages. 
\paragraph{Instruction-following.} We further evaluated the instruction-following capabilities of the generated models, as the ability to accurately interpret and execute user prompts in practical scenarios constitutes an essential aspect of model quality. To this end, we employed the IFEval benchmark~\cite{zhou2023instructionfollowingevaluationlargelanguage}, which is specifically designed to automatically measure instruction adherence across a diverse set of tasks, including the inclusion or exclusion of keywords, the use of certain punctuation marks and capitalization patterns, and similar formal requirements. 
IFEval was originally proposed for English, and subsequent work introduced translated and post-edited versions for Catalan\footnote{\url{https://huggingface.co/datasets/projecte-aina/IFEval_ca}} and Spanish.\footnote{\url{https://huggingface.co/datasets/BSC-LT/IFEval_es}} As part of this study, we extended the benchmark to Basque using GPT-4o, followed by manual revision by native speakers to ensure semantic fidelity, fluency, and naturalness. Instances that did not translate directly due to linguistic or cultural differences were adapted accordingly. The metadata and IFEval codebase were also modified to ensure correct evaluation in each of the supported languages. Note that we excluded the task type \texttt{response\_language} from our evaluation (5\% of the instances), due to discrepancies across the language-specific versions of the dataset.

\begin{table}[]
    \centering
    \resizebox{\linewidth}{!}{
    \begin{tabular}{l|ccccc}
        \toprule
        \textbf{Model} & \textbf{\texttt{EU}} & \textbf{\texttt{GL}} & \textbf{\texttt{CA}} & \textbf{\texttt{ES}} & \textbf{\texttt{EN}} \\ \midrule
        Llama 3.1 8B & $47.37$ & $58.91$ & $59.14$ & $\mathbf{66.36}$ & $72.80$ \\
        Llama 3.1 8B\textsubscript{ eu} & $\mathbf{60.63}$ & $56.19$ & $56.77$ & $64.37$ & $\mathbf{74.85}$ \\
        Llama 3.1 8B\textsubscript{ gl} & $43.94$ & $\mathbf{59.59}$ & $56.89$ & $63.44$ & $71.08$ \\
        Llama 3.1 8B\textsubscript{ ca} & $44.13$ & $54.64$ & $\mathbf{60.91}$ & $64.64$ & $70.91$ \\
        Llama 3.1 8B\textsubscript{ es} & $45.51$ & $57.81$ & $58.56$ & $65.88$ & $72.13$ \\
        \midrule
        Qwen3 8B & $51.02$ & $61.80$ & $62.81$ & $\mathbf{68.57}$ & $75.27$ \\
        Qwen3 8B\textsubscript{ eu} & $\mathbf{67.56}$ & $60.47$ & $60.72$ & $66.13$ & $\mathbf{78.30}$ \\
        Qwen3 8B\textsubscript{ gl} & $48.66$ & $\mathbf{62.57}$ & $61.76$ & $66.09$ & $74.47$ \\
        Qwen3 8B\textsubscript{ ca} & $49.63$ & $59.56$ & $\mathbf{64.60}$ & $66.78$ & $74.79$ \\
        Qwen3 8B\textsubscript{ es} & $50.79$ & $62.24$ & $63.57$ & $67.11$ & $75.58$ \\
        \midrule
        Qwen3 14B & $55.57$ & $63.37$ & $66.55$ & $\mathbf{70.98}$ & $77.06$ \\
        Qwen3 14B\textsubscript{ eu} & $\mathbf{70.77}$ & $61.10$ & $62.86$ & $69.52$ & $\mathbf{80.36}$ \\
        Qwen3 14B\textsubscript{ gl} & $53.66$ & $\mathbf{66.09}$ & $64.83$ & $68.89$ & $77.05$ \\
        Qwen3 14B\textsubscript{ ca} & $54.22$ & $61.09$ & $\mathbf{68.06}$ & $69.37$ & $77.19$ \\
        Qwen3 14B\textsubscript{ es} & $55.90$ & $63.62$ & $65.91$ & $70.48$ & $77.51$ \\
        \bottomrule
    \end{tabular}
    }
    \caption{Base model language adaptation results. Bold indicates best among the same model architecture and underline indicates best overall.}
    \label{tab:base-model-results}
\end{table}

\begin{table*}[!ht]
    \centering
    \resizebox{\textwidth}{!}{
    \begin{tabular}{l|ccccc|ccccc}
        \toprule 
          & \multicolumn{5}{c|}{\textbf{Benchmark average}} & \multicolumn{5}{c}{\textbf{Machine Translation}} \\
         \textbf{Model} & \textbf{\texttt{EU}} & \textbf{\texttt{GL}} & \textbf{\texttt{CA}} & \textbf{\texttt{ES}} & \textbf{\texttt{EN}} & \textbf{\texttt{*-EU}} & \textbf{\texttt{*-GL}} & \textbf{\texttt{*-CA}} & \textbf{\texttt{*-ES}} & \textbf{\texttt{*-EN}} \\ \midrule
    
         Llama 3.1 8B\textsubscript{ joint-EU} & $61.75$ & $58.13$ & $57.81$ & $64.59$ & $73.71$ & $15.03$ & $25.71$ & $29.00$ & $23.86$ & $35.42$ \\
         Salamandra 2B\textsubscript{ Instruct} & $27.95$ & $37.11$ & $43.18$ & $34.68$ & $37.13$ & $6.69$ & $25.27$ & $28.41$ & $21.69$ & $31.50$ \\
         Salamandra 7B\textsubscript{ Instruct} & $44.94$ & $53.60$ & $56.55$ & $52.79$ & $57.27$ & $11.31$ & $\underline{29.94}$ & $\underline{34.12}$ & $25.46$ & $37.67$ \\
         ALIA 40B\textsubscript{ Instruct} & $60.64$ & $64.98$ & $64.68$ & $62.93$ & $66.04$ & $\underline{15.78}$ & $29.86$ & $33.10$ & $\underline{26.56}$ & $\underline{37.85}$ \\ \midrule
         Llama 3.1 8B\textsubscript{ Instruct} &  $49.29$ & $60.11$ & $61.56$ & $68.22$ & $73.87$ & $7.18$ & $26.55$ & $30.12$ & $24.01$ & $35.49$ \\
         Llama 3.1 8B\textsubscript{ merge-EU} & $\mathbf{58.36}$ & $61.41$ & $60.61$ & $67.87$ & $\mathbf{75.08}$ & $\mathbf{12.27}$ & $26.14$ & $29.91$ & $23.97$ & $\mathbf{37.46}$  \\
         Llama 3.1 8B\textsubscript{ merge-GL} & $47.56$ & $\mathbf{63.94}$ & $59.96$ & $68.23$ &$73.56$ & $6.37$ & $\mathbf{28.91}$ & $27.81$ & $23.76$ & $36.69$ \\
         Llama 3.1 8B\textsubscript{ merge-CA} & $48.26$ & $59.84$ & $\mathbf{63.99}$ & $67.87$ & $73.48$ & $6.85$ & $24.49$ & $\mathbf{32.92}$ & $23.79$ & $36.77$ \\
         Llama 3.1 8B\textsubscript{ merge-ES} & $40.46$ & $60.26$ & $60.62$ & $\mathbf{68.87}$ & $74.24$ & $7.65$ & $26.91$ & $30.84$ & $\mathbf{24.46}$ & $36.71$ \\
         Llama 3.1 8B\textsubscript{ merge-multi} & $51.66$ & $62.23$ & $61.72$ & $68.46$ & $74.06$ & $8.42$ & $27.19$ & $31.10$ & $24.31$ & $37.02$ \\ \midrule
    
         Qwen3 8B\textsubscript{ Instruct} & $44.06$ & $56.51$ & $59.37$ & $64.04$ & $69.84$ & $3.51$ & $24.39$ & $27.78$ & $23.05$ & $33.09$ \\
         Qwen3 8B\textsubscript{ merge-EU} & $\mathbf{55.81}$ & $55.80$ & $60.94$ & $65.80$ & $\mathbf{72.67}$ & $\mathbf{9.39}$ & $24.60$ & $27.64$ & $\mathbf{24.06}$ & $\mathbf{34.57}$ \\
         Qwen3 8B\textsubscript{ merge-GL} & $38.98$ & $\mathbf{58.84}$ & $59.77$ & $65.22$ & $71.47$ & $2.74$ & $\mathbf{27.35}$ & $25.22$ & $23.37$ & $33.19$ \\
         Qwen3 8B\textsubscript{ merge-CA} & $39.29$ & $56.27$ & $\mathbf{62.66}$ & $65.90$ & $71.10$ & $3.11$ & $24.01$ & $\mathbf{31.29}$ & $23.46$ & $33.78$ \\
         Qwen3 8B\textsubscript{ merge-ES} & $41.03$ & $56.74$ & $60.26$ & $\mathbf{65.91}$ & $72.44$ & $3.49$ & $25.00$ & $28.00$ & $23.90$ & $33.32$ \\
         Qwen3 8B\textsubscript{ merge-multi} & $44.85$ & $56.64$ & $61.34$ & $65.65$ & $72.19$ & $4.45$ & $26.02$ & $28.83$ & $23.80$ & $33.86$ \\ \midrule
    
         Qwen3 14B\textsubscript{ Instruct} & $52.09$ & $62.39$ & $62.14$ & $67.87$ & $71.82$ & $5.40$ & $25.87$ & $29.11$ & $24.20$ & $35.10$ \\
         Qwen3 14B\textsubscript{ merge-EU} & $\underline{\mathbf{63.26}}$ & $62.66$ & $64.62$ & $68.88$ & $\underline{\mathbf{75.78}}$ & $\mathbf{11.78}$ & $26.45$ & $29.99$ & $\mathbf{25.19}$ & $\mathbf{36.86}$ \\
         Qwen3 14B\textsubscript{ merge-GL} & $53.96$ & $\underline{\mathbf{66.27}}$ & $64.46$ & $69.30$ & $74.28$ & $5.46$ & $\mathbf{29.21}$ & $29.52$ & $24.55$ & $36.10$ \\
         Qwen3 14B\textsubscript{ merge-CA} & $53.85$ & $63.28$ & $\underline{\mathbf{66.84}}$ & $69.39$ & $74.50$ & $5.49$ & $25.92$ & $\mathbf{32.40}$ & $24.54$ & $36.22$ \\
         Qwen3 14B\textsubscript{ merge-ES} & $53.52$ & $64.02$ & $64.90$ & $\underline{\mathbf{69.98}}$ & $74.78$ & $5.86$ & $27.04$ & $30.02$ & $24.78$ & $36.21$ \\
         Qwen3 14B\textsubscript{ merge-multi} & $56.37$ & $64.15$ & $65.17$ & $69.47$ &$74.77$ & $7.14$ & $27.63$ & $30.91$ & $24.88$ & $36.45$ \\
         \bottomrule
    \end{tabular}
    
    }
    \caption{Main experimental results on multiple-choice benchmarks (Accuracy) and machine translation (BLEU). Bold indicates best among the same backbone model and underline indicates best overall.}
    \label{tab:tabla-linear}
    \vspace{-.4em}
\end{table*}

\section{Results} \label{sec:results}

In this section, we first discuss the main findings (§~\ref{ssec:main-results}), beginning with the performance of the language-adapted base models, followed by the monolingual and multilingual merging strategies, and finally their instruction-following capabilities. We then report additional experimental analyses (§~\ref{ssec:further-analyses}), examining the impact of different merging methods and merge proportions.

The results are organized in a top-down manner. We start by comparing the models obtained using the best-performing merging approach, \textit{Linear}, with the best hyperparameters ($w_i=1.$ for monolingual merging and $w_i=.25$ for multilingual merging), and subsequently present the development analyses in greater detail.

\subsection{Main results} \label{ssec:main-results}


\paragraph{Base models results.} \label{par:base-results}

Table~\ref{tab:base-model-results} compares the language-adapted base models against their corresponding non-adapted base baselines on multiple-choice benchmarks. Overall, the language-adapted versions \textbf{clearly outperform the baseline in their target languages}, with the exception of the Spanish, an already high-resource language.
The biggest improvements are found in Basque, which is expected due to the low amount of resources and the lower baseline results. We can conclude that the language-adapted models achieve consistently better performance than the original base models and, therefore, can be used to teach the target languages to the instructed variants.

\paragraph{Monolingual merge results.} \label{par:monolingual-results}

Table~\ref{tab:tabla-linear} compares the performance of language-specific merged models (\texttt{merge-}) with the original instruct models (\texttt{instruct}) across multiple-choice benchmarks and machine translation tasks. The results show that monolingual merges generally improve performance in their target language relative to the non-adapted instruct baseline (\texttt{merge-} vs.\ \texttt{instruct}), with the clearest gains observed for lower-resource languages (EU/GL/CA), while results for Spanish are more mixed. In machine translation, monolingual merged models also tend to improve translation quality in their target language compared to the instruct baselines, mirroring the trends observed in the benchmark evaluation. When compared with the state-of-the-art language adaptation approach (Llama 3.1 8B\textsubscript{joint-EU}), the Basque merged models achieve comparable performance in the target language across both benchmarks and translation tasks, while exhibiting substantially smaller performance degradation in other languages, resulting in more consistent and stable overall performance and highlighting the promise of the merging approach. Finally, similarly strong improvements are observed across the different backbone models, further demonstrating the robustness of the model-merging strategy.

\paragraph{Multilingual results.} \label{par:multilingual-results}

Table~\ref{tab:tabla-linear} also reports the results for the multilingual merged models (listed as \texttt{{model}}\textsubscript{merge-multi}). Compared with the monolingual merged variants, the multilingual models generally provide more balanced performance across languages, although their performance typically remains below that of the target-language monolingual models. Nevertheless, relative to the instruct baseline, they show clear improvements in almost every language across all backbone models. Overall, these results indicate that (1) model merging is a promising approach for enabling an instructed LLM to acquire competence in multiple languages, and (2) there remains significant room for improvement in transferring language competence between models.

\begin{table}[t]
    \resizebox{\linewidth}{!}{
    \begin{tabular}{l|c@{\hspace{1.5pt}}lc@{\hspace{1.5pt}}lc@{\hspace{1.5pt}}lc@{\hspace{1.5pt}}l}
        \toprule
        \textbf{Model}                           
            & \multicolumn{2}{c}{\textbf{\texttt{EU}}} 
            & \multicolumn{2}{c}{\textbf{\texttt{GL}}} 
            & \multicolumn{2}{c}{\textbf{\texttt{CA}}} 
            & \multicolumn{2}{c}{\textbf{\texttt{ES}}}\\
        \midrule
        Llama 3.1 8B\textsubscript{ joint-EU}    & $46.82$   & \scriptsize $ 2.0$   & $48.36$   & \scriptsize $ 2.6$   & $46.71$   & \scriptsize $ 0.7$   & $57.06$   & \scriptsize $ 0.2$ \\
        Salamandra 2B\textsubscript{ Instruct}   & $23.87$   & \scriptsize $ 0.9$   & $24.78$   & \scriptsize $ 0.7$   & $25.20$   & \scriptsize $ 0.7$   & $25.15$   & \scriptsize $ 0.9$ \\
        Salamandra 7B\textsubscript{ Instruct}   & $32.79$   & \scriptsize $ 1.5$   & $35.12$   & \scriptsize $ 0.7$   & $34.93$   & \scriptsize $ 2.8$   & $33.50$   & \scriptsize $ 2.0$ \\
        ALIA 40B\textsubscript{ Instruct}        & $35.41$   & \scriptsize $ 0.3$   & $38.40$   & \scriptsize $ 0.7$   & $41.17$   & \scriptsize $ 0.4$   & $43.48$   & \scriptsize $ 1.6$ \\
        \midrule
        Llama 3.1 8B\textsubscript{ Instruct}    & $\mathbf{40.22}$   & \scriptsize $ 1.2$   & $\mathbf{58.66}$   & \scriptsize $ 1.7$   & $\mathbf{52.03}$   & \scriptsize $ 0.7$   & $\mathbf{63.83}$   & \scriptsize $ 0.9$ \\
        Llama 3.1 8B\textsubscript{ merge-*} & $39.98$   & \scriptsize $ 1.1$   & $47.86$   & \scriptsize $ 1.3$   & $46.96$   & \scriptsize $ 1.2$   & $50.08$   & \scriptsize $ 0.7$ \\
        Llama 3.1 8B\textsubscript{ merge-multi} & $34.08$   & \scriptsize $ 1.4$   & $47.74$   & \scriptsize $ 1.0$   & $45.03$   & \scriptsize $ 1.9$   & $51.59$   & \scriptsize $ 0.4$ \\
        \midrule
        Qwen3 8B\textsubscript{ Instruct}        & $\mathbf{54.05}$   & \scriptsize $ 2.1$   & $\mathbf{75.97}$   & \scriptsize $ 0.5$   & $\mathbf{74.59}$   & \scriptsize $ 0.9$   & $\mathbf{80.66}$   & \scriptsize $ 0.9$ \\
        Qwen3 8B\textsubscript{ merge-*}    & $43.38$   & \scriptsize $ 1.1$   & $63.06$   & \scriptsize $0.1$   & $59.45$   & \scriptsize $0.8$   & $66.83$   & \scriptsize $ 0.2$ \\
        Qwen3 8B\textsubscript{ merge-multi}     & $39.98$   & \scriptsize $ 1.6$   & $62.85$   & \scriptsize $ 0.4$   & $61.51$   & \scriptsize $ 0.7$   & $66.75$   & \scriptsize $ 1.1$ \\
        \midrule 
        Qwen3 14B\textsubscript{ Instruct}       & $\mathbf{62.81}$   & \scriptsize $ 0.9$   & $\mathbf{79.00}$   & \scriptsize $ 0.9$   & $\mathbf{76.81}$   & \scriptsize $ 0.3$   & $\mathbf{82.92}$   & \scriptsize $ 0.4$ \\
        Qwen3 14B\textsubscript{ merge-*}   & $61.35$   & \scriptsize $ 0.6$   & $65.75$   & \scriptsize $ 1.7$   & $66.00$   & \scriptsize $ 1.1$   & $71.43$   & \scriptsize $ 1.1$ \\
        Qwen3 14B\textsubscript{ merge-multi}    & $56.45$  & \scriptsize $ 1.0$   & $66.25$  & \scriptsize $ 1.0$   & $67.25$   & \scriptsize $ 1.9$   & $72.26$   & \scriptsize $ 0.9$ \\
        \bottomrule
    \end{tabular}
    }
    \caption{Strict instruction-level accuracy on IFEval (mean $\pm$ SD over 3 runs). For each test language, \texttt{\{model\}}\textsubscript{merge-*} denotes the corresponding monolingual variant adapted to that language.}
    \label{tab:ifeval-results}
    \vspace{-.2em}
\end{table}

\paragraph{Instruction-following results.} \label{par:inst-following-results}

Table~\ref{tab:ifeval-results} presents the instruction-following performance across Basque, Galician, Catalan, and Spanish. In contrast to the consistent gains observed in static benchmarks and translation tasks, model merging yields rather mixed results in this setting. These findings can be interpreted from two complementary perspectives. On the one hand, both monolingual and multilingual merges retain instruction-following capabilities to some extent, suggesting that the merging process successfully transfers these abilities to the language-adapted base models alongside the improvements observed in the benchmarks. On the other hand, the fact that the merged models do not surpass the instruct baseline indicates that improvements in language proficiency do not necessarily translate into stronger instruction-following capabilities when using the merging approach. The gap becomes even more evident when compared with the state-of-the-art joint adaptation method. Nevertheless, as discussed later in \S\ref{ssec:further-analyses}, instruction-following performance varies substantially depending on the merging method used.

\begin{table}[t]
    \centering
    \resizebox{\linewidth}{!}{
    \begin{tabular}{l|cccc|c}
        \toprule
        \textbf{Method} & \textbf{\texttt{EU}} & \textbf{\texttt{GL}} & \textbf{\texttt{CA}} & \textbf{\texttt{ES}} & \textbf{\texttt{Avg}} \\\midrule
        \multicolumn{6}{c}{Benchmark average} \\ \midrule
         Linear           & $58.36$          & $\mathbf{63.94}$ & $\mathbf{63.99}$ & $\mathbf{68.87}$ & $\mathbf{63.79}$ \\
         Task Arithmetic  & $61.48$          & $61.67$          & $63.62$          & $67.61$          & $63.59$ \\
         DARE             & $61.23$          & $61.55$          & $63.70$          & $66.29$          & $63.19$ \\
         Breadcrumbs      & $\mathbf{61.53}$ & $61.22$          & $63.89$          & $67.54$          & $63.55$ \\
        \midrule
        \multicolumn{6}{c}{Machine translation} \\ \midrule
         Linear           & $12.27$          & $\mathbf{28.91}$ & $\mathbf{32.92}$ & $\mathbf{24.46}$ & $\mathbf{24.64}$ \\
         Task Arithmetic  & $\mathbf{13.62}$ & $28.26$          & $32.76$          & $23.78$          & $24.61$ \\
         DARE             & $12.35$          & $27.90$          & $32.65$          & $23.39$          & $24.07$ \\
         Breadcrumbs      & $13.35$          & $00.19$          & $00.42$          & $22.46$          & $09.11$ \\
         \midrule
         \multicolumn{6}{c}{Instruction Following} \\ \midrule
         Linear           & $39.98$          & $\mathbf{47.86}$          & $\mathbf{46.96}$          & $50.08$          & $\mathbf{46.22}$ \\
         Task Arithmetic  & $50.73$          & $35.04$          & $42.10$          & $\mathbf{54.59}$          & $45.62$ \\
         DARE             & $\mathbf{50.77}$          & $35.33$          & $42.10$          & $23.56$          & $37.94$ \\
         Breadcrumbs      & $46.45$          & $29.93$          & $28.85$          & $51.15$          & $39.09$ \\
        \bottomrule
    \end{tabular}
    }
    \caption{Merge method comparison. Results are reported using Llama 3.1 8B\textsubscript{ merge-*}, where each language is evaluated with their specialized expert.}
    \label{tab:method-comparison}
    \vspace{-.2em}
\end{table}


\begin{figure*}[t]
  \centering
  \includegraphics[width=\linewidth]{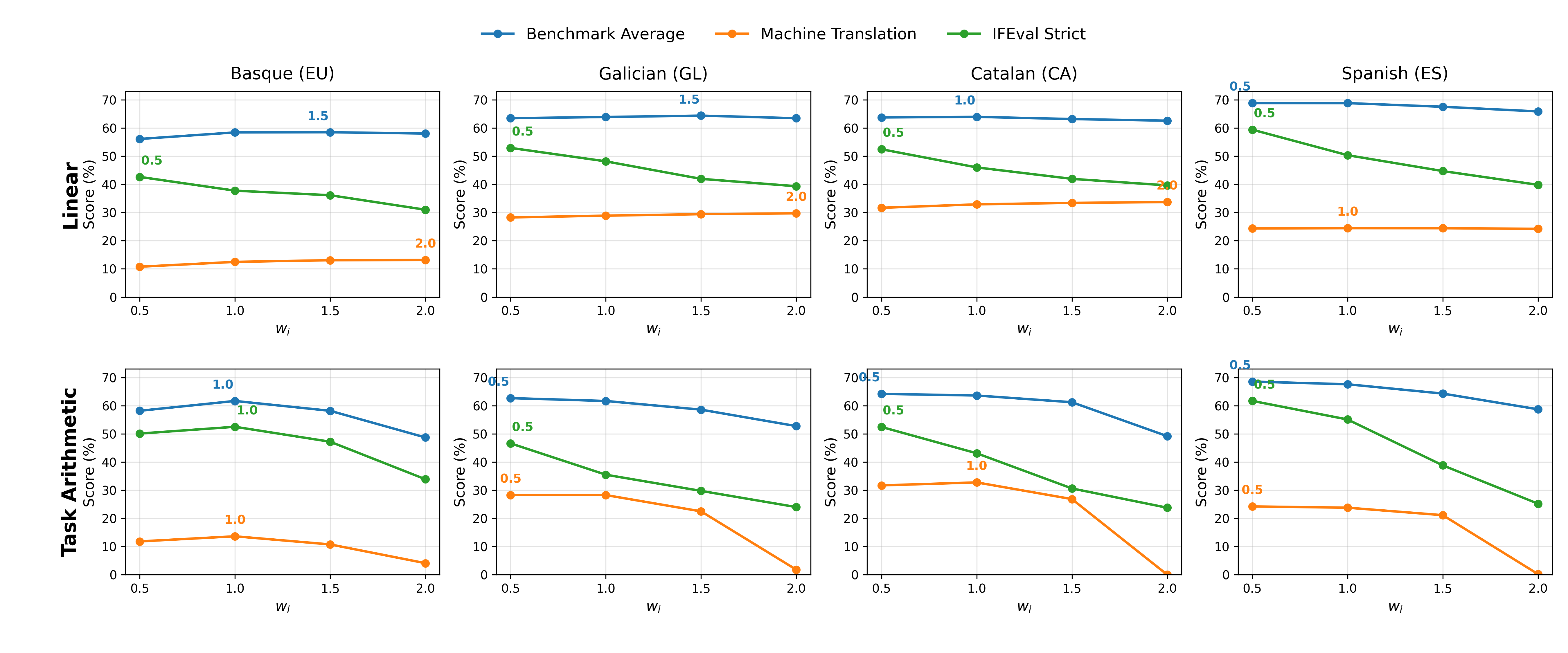}
  \caption{Proportion sweep ablation for Llama 3.1 8B merges across Iberian languages (EU, GL, CA, ES), comparing Linear (top row) and Task Arithmetic (bottom row). Each subplot shows the effect of varying the merge proportion $w_i$ on Benchmark Average, Machine Translation, and IFEval Strict.}
  \label{fig:proportion-sweeps}
\end{figure*}

\subsection{Further analyses results} \label{ssec:further-analyses} 

Beyond the main experiments, this section provides complementary analyses that investigate key preliminary factors influencing our results. We first examine the effect of the various merging strategies employed, followed by an analysis of the influence of the weighting parameter $w_i$ in the context of monolingual merges.

\paragraph{Merge method comparison.}

\begin{table}[t]
    \centering
    \resizebox{\linewidth}{!}{
    \begin{tabular}{l|cccc|c}
        \toprule
        \textbf{Method} & \textbf{\texttt{EU}} & \textbf{\texttt{GL}} & \textbf{\texttt{CA}} & \textbf{\texttt{ES}} & \textbf{\texttt{Avg}} \\ \midrule
        \multicolumn{6}{c}{Benchmark average} \\ \midrule
         Linear           & $55.81$          & $58.84$          & $\mathbf{62.66}$ & $\mathbf{65.91}$ & $\mathbf{60.80}$ \\
         Task Arithmetic  & $55.71$          & $57.26$          & $58.14$          & $63.55$          & $58.66$ \\
         DARE             & $52.39$          & $55.97$          & $55.42$          & $61.94$          & $56.43$ \\
         Breadcrumbs      & $\mathbf{56.38}$ & $\mathbf{59.35}$ & $61.33$          & $65.41$          & $60.62$ \\
         \midrule
         \multicolumn{6}{c}{Machine translation} \\ \midrule
         Linear           & $9.39$           & $27.35$          & $31.29$          & $\mathbf{23.90}$ & $22.98$ \\
         Task Arithmetic  & $\mathbf{12.65}$ & $\mathbf{29.27}$ & $\mathbf{32.23}$ & $23.36$          & $\mathbf{24.38}$ \\
         DARE             & $12.22$          & $28.89$          & $32.09$          & $23.22$          & $24.11$ \\
         Breadcrumbs      & $0.11$           & $3.86$           & $0.61$           & $22.55$          & $6.78$ \\
         \midrule
         \multicolumn{6}{c}{Instruction Following} \\ \midrule
         Linear           & $43.38$          & $63.06$          & $59.45$          & $66.83$          & $58.18$ \\
         Task Arithmetic  & $55.92$          & $\mathbf{76.21}$ & $\mathbf{77.69}$ & $81.04$          & $72.72$ \\
         DARE             & $\mathbf{61.02}$ & $75.26$          & $77.19$          & $\mathbf{81.08}$ & $\mathbf{73.64}$ \\
         Breadcrumbs      & $27.69$          & $35.95$          & $32.12$          & $51.92$          & $36.92$ \\
        \bottomrule
    \end{tabular}
    }
    \caption{Merge method comparison. Results are reported using Qwen3 8B\textsubscript{ merge-*}, where each language is evaluated with their specialized expert.}
    \label{tab:method-comparison-qwen}
    \vspace{-.2em}
\end{table}

During our experimental phase, we evaluated the various merging methods described in \S\ref{ssec:merge-techniques}. Table~\ref{tab:method-comparison} presents the results obtained by each method for each language with its corresponding specialized expert across three scenarios: benchmarks, machine translation, and instruction following. Overall, although the \textit{Linear} method appears to perform best across most language–task pairs, \textbf{there is no clear one-size-fits-all method}. For example, for the lowest-resource language (Basque), the \textit{Linear} method is the worst-performing approach, while the remaining methods achieve comparable performance. For the machine translation task, where text generation is required, we observe that the \textit{Breadcrumbs} method performs particularly poorly, leading to a complete collapse of the model for Galician and Catalan. Finally, instruction-following capabilities exhibit the greatest variability across methods, with no clear pattern indicating which method performs best. Moreover, after conducting the same analysis on the other models, see Table~\ref{tab:method-comparison-qwen}, we conclude that \textbf{there is no universally best method}; instead, performance of the merging method must be assessed on a case-by-case basis for each language and model pair.

\paragraph{Merge proportion impact.}

After evaluating the impact of different merging methods, we next explored the effect of the merge proportion \(w_i\). For this analysis, we restricted the setup to two methods: \textit{Linear} and \textit{Task Arithmetic}. These were selected because they showed the strongest overall performance in the previous analysis. The \textit{Linear} method represents the simplest interpolation approach, while \textit{Task Arithmetic} serves as the baseline for several more advanced merging techniques. Figure~\ref{fig:proportion-sweeps} presents the proportion sweep across the four Iberian languages (Basque, Galician, Catalan, and Spanish), reporting Benchmark Average, Machine Translation, and Instruction Following performance. The best \(w_i\) value for each configuration is annotated in the figure.

As expected, for both approaches, increasing \(w_i\) increases the influence of the language-adapted base model and therefore tends to degrade instruction-following capabilities. However, the behavior differs across the other two axes, Benchmark Average and Machine Translation. For the \textit{Linear} method, the trade-off is relatively smooth: performance on benchmarks and machine translation does not exhibit a sharp degradation as \(w_i\) increases. In contrast, with \textit{Task Arithmetic}, the model rapidly collapses once \(w_i\) exceeds a certain threshold (typically \(w_i \geq 1.5\)). Interestingly, the effect of \(w_i\) on benchmark performance and machine translation appears to be strongly correlated. Finally, although the optimal \(w_i\) value varies across languages, the general trends remain consistent: selecting a value in the range \(w_i \in [0.5, 1.0)\) typically yields robust performance.

\section{Conclusions}

In this work, we show that model merging is a feasible alternative to continual pre-training for extending instructed LLMs to low-resource languages. Our experiments across Basque, Galician, Catalan, and Spanish show that merging can successfully transfer the target language proficiency from specialized base models into instructed variants. As a result, the obtained language-adapted and instruction-following models show substantial performance gains on benchmarks and machine translation tasks while maintaining instruction following capabilities, particularly for under-represented languages such as Basque. Among the methods evaluated, Linear and Task Arithmetic have shown to be the best performing alternatives overall, confirming that even simple parameter-space merging strategies can effectively ``teach'' new languages to instruction-tuned models. 

Beyond raw performance, merging offers an efficient path for multilingual expansion that avoids the heavy computational cost of re-training and fine-tuning cycles. The approach lowers hardware requirements, enabling smaller research groups to adapt frontier models to their own languages. Still, instruction-following abilities remain partly sensitive to merging, emphasizing the need for techniques that preserve alignment during language transfer. Future work should explore merge alignment, investigate stability across model families, and design language-specific merging algorithms to further improve instruction retention. 

Overall, our findings suggest that model merging bridges the gap between efficiency and multilingual coverage, providing a promising direction for building more accessible language models. As part of the contributions of this work, we publicly release the continued pre-trained base models and the Basque and Galician IFEval variants.\footnote{https://huggingface.co/collections/HiTZ/merge-and-conquer}

\section{Limitations}

Despite the promising results, this work has several limitations. Our study is restricted to a limited number of model families and sizes of up to 14B parameters, as well as to a small set of Iberian languages---namely Basque, Catalan, Galician, and Spanish. Extending these experiments to larger model scales and additional languages would help to further validate and generalize our findings. Moreover, our analysis focuses exclusively on instruction-tuned models, while model merging techniques could also be valuable in other contexts, such as judge LLMs, reward models, or safety and guard models.

\section*{Acknowledgments}

This work has been funded by the Spanish Ministry of Science, Innovation, and Universities (Project HumanAIze, grant number AIA2025-163322-C61), by the Basque Government (IKER-GAITU project) and the Ministerio para la Transformación Digital y de la Función Pública - Funded by EU – NextGenerationEU within the framework of the project Desarrollo de Modelos ALIA. The models were trained on the Leonardo supercomputer at CINECA under the EuroHPC Joint Undertaking, project EHPC-EXT-2024E01-042. We also acknowledge the support of the HiTZ Chair of Artificial Intelligence and Language Technology (TSI100923-2023-1), funded by MTDFP, Secretaría de Estado de Digitalización e Inteligencia Artificial, ENIA, and by the European Union-Next Generation EU / PRTR. We thank Sofía García González for post-editing the Galician translation of IFEval.

\nocite{*}

\section{Bibliographical References}
\label{sec:reference}
\bibliographystyle{lrec2026-natbib}
\bibliography{lrec2026-example}

\begin{thebibliography}{53}
\expandafter\ifx\csname natexlab\endcsname\relax\def\natexlab#1{#1}\fi

\bibitem[{Akiba et~al.(2025)Akiba, Shing, Tang, Sun, and Ha}]{Akiba2024EvolutionaryOOC}
Takuya Akiba, Makoto Shing, Yujin Tang, Qi~Sun, and David Ha. 2025.
\newblock \href {https://doi.org/10.1038/s42256-024-00975-8} {Evolutionary optimization of model merging recipes}.
\newblock \emph{Nature Machine Intelligence}, 7(2):195--204.

\bibitem[{Bai et~al.(2023)Bai, Bai, Chu, Cui, Dang, Deng, Fan, Ge, Han, Huang, Hui, Ji, Li, Lin, Lin, Liu, Liu, Lu, Lu, Ma, Men, Ren, Ren, Tan, Tan, Tu, Wang, Wang, Wang, Wu, Xu, Xu, Yang, Yang, Yang, Yang, Yao, Yu, Yuan, Yuan, Zhang, Zhang, Zhang, Zhang, Zhou, Zhou, Zhou, and Zhu}]{bai2023qwentechnicalreport}
Jinze Bai, Shuai Bai, Yunfei Chu, Zeyu Cui, Kai Dang, Xiaodong Deng, Yang Fan, Wenbin Ge, Yu~Han, Fei Huang, Binyuan Hui, Luo Ji, Mei Li, Junyang Lin, Runji Lin, Dayiheng Liu, Gao Liu, Chengqiang Lu, Keming Lu, Jianxin Ma, Rui Men, Xingzhang Ren, Xuancheng Ren, Chuanqi Tan, Sinan Tan, Jianhong Tu, Peng Wang, Shijie Wang, Wei Wang, Shengguang Wu, Benfeng Xu, Jin Xu, An~Yang, Hao Yang, Jian Yang, Shusheng Yang, Yang Yao, Bowen Yu, Hongyi Yuan, Zheng Yuan, Jianwei Zhang, Xingxuan Zhang, Yichang Zhang, Zhenru Zhang, Chang Zhou, Jingren Zhou, Xiaohuan Zhou, and Tianhang Zhu. 2023.
\newblock \href {https://qwenlm.github.io/blog/qwen/} {Qwen technical report}.
\newblock Technical report.

\bibitem[{Bandarkar et~al.(2024)Bandarkar, Liang, Muller, Artetxe, Shukla, Husa, Goyal, Krishnan, Zettlemoyer, and Khabsa}]{bandarkar-etal-2024-belebele}
Lucas Bandarkar, Davis Liang, Benjamin Muller, Mikel Artetxe, Satya~Narayan Shukla, Donald Husa, Naman Goyal, Abhinandan Krishnan, Luke Zettlemoyer, and Madian Khabsa. 2024.
\newblock \href {https://doi.org/10.18653/v1/2024.acl-long.44} {The belebele benchmark: a parallel reading comprehension dataset in 122 language variants}.
\newblock In \emph{Proceedings of the 62nd Annual Meeting of the Association for Computational Linguistics (Volume 1: Long Papers)}, pages 749--775, Bangkok, Thailand. Association for Computational Linguistics.

\bibitem[{Bao et~al.(2023)Bao, Perez, and Parapar}]{Bao2023ConversationsIGC}
Eliseo Bao, Anxo Perez, and Javier Parapar. 2023.
\newblock \href {https://api.semanticscholar.org/CorpusId:265043707} {Conversations in galician: a large language model for an underrepresented language}.
\newblock \emph{ArXiv}, abs/2311.03812.

\bibitem[{Baucells et~al.(2025)Baucells, Aula-Blasco, de~Dios-Flores, Paniagua~Su{\'a}rez, Perez, Salles, Sotelo~Docio, Falc{\~a}o, Saiz, Sepulveda~Torres, Barnes, Gamallo, Gonzalez-Agirre, Rigau, and Villegas}]{baucells-etal-2025-iberobench}
Irene Baucells, Javier Aula-Blasco, Iria de~Dios-Flores, Silvia Paniagua~Su{\'a}rez, Naiara Perez, Anna Salles, Susana Sotelo~Docio, J{\'u}lia Falc{\~a}o, Jose~Javier Saiz, Robiert Sepulveda~Torres, Jeremy Barnes, Pablo Gamallo, Aitor Gonzalez-Agirre, German Rigau, and Marta Villegas. 2025.
\newblock \href {https://aclanthology.org/2025.coling-main.699/} {{I}bero{B}ench: A benchmark for {LLM} evaluation in {I}berian languages}.
\newblock In \emph{Proceedings of the 31st International Conference on Computational Linguistics}, pages 10491--10519, Abu Dhabi, UAE. Association for Computational Linguistics.

\bibitem[{Bel et~al.(2024{\natexlab{a}})Bel, Punsola, and Ruiz-Fern{\'a}ndez}]{bel-etal-2024-escola}
N{\'u}ria Bel, Marta Punsola, and Valle Ruiz-Fern{\'a}ndez. 2024{\natexlab{a}}.
\newblock \href {https://aclanthology.org/2024.lrec-main.554/} {{E}s{C}o{LA}: {S}panish corpus of linguistic acceptability}.
\newblock In \emph{Proceedings of the 2024 Joint International Conference on Computational Linguistics, Language Resources and Evaluation (LREC-COLING 2024)}, pages 6268--6277, Torino, Italia. ELRA and ICCL.

\bibitem[{Bel et~al.(2024{\natexlab{b}})Bel, Punsola, and Ruiz-Fernández}]{PLN6609}
Núria Bel, Marta Punsola, and Valle Ruiz-Fernández. 2024{\natexlab{b}}.
\newblock \href {http://journal.sepln.org/sepln/ojs/ojs/index.php/pln/article/view/6609} {Catcola, catalan corpus of linguistic acceptability}.
\newblock \emph{Procesamiento del Lenguaje Natural}, 73(0):177--190.

\bibitem[{Cao et~al.(2025)Cao, Wu, Prasad, Tian, and Liu}]{cao2025paramdelta}
Sheng Cao, Mingrui Wu, Karthik Prasad, Yuandong Tian, and Zechun Liu. 2025.
\newblock \href {https://openreview.net/forum?id=vqbd2OQnGp} {Param{$\Delta$} for direct mixing: Post-train large language model at zero cost}.
\newblock In \emph{The Thirteenth International Conference on Learning Representations}.
\newblock ICLR 2025 poster.

\bibitem[{Davari and Belilovsky(2025)}]{10.1007/978-3-031-73226-3_16}
MohammadReza Davari and Eugene Belilovsky. 2025.
\newblock Model breadcrumbs: Scaling multi-task model merging with sparse masks.
\newblock In \emph{Computer Vision -- ECCV 2024}, pages 270--287, Cham. Springer Nature Switzerland.

\bibitem[{de~Dios-Flores et~al.(2024)de~Dios-Flores, Su{\'a}rez, P{\'e}rez, Outeiri{\~n}o, Garcia, and Gamallo}]{de-dios-flores-etal-2024-corpusnos}
Iria de~Dios-Flores, Silvia~Paniagua Su{\'a}rez, Cristina~Carbajal P{\'e}rez, Daniel~Bardanca Outeiri{\~n}o, Marcos Garcia, and Pablo Gamallo. 2024.
\newblock \href {https://aclanthology.org/2024.propor-1.66/} {{C}orpus{N{\'O}S}: A massive {G}alician corpus for training large language models}.
\newblock In \emph{Proceedings of the 16th International Conference on Computational Processing of Portuguese - Vol. 1}, pages 593--599, Santiago de Compostela, Galicia/Spain. Association for Computational Lingustics.

\bibitem[{Elhady et~al.(2025)Elhady, Agirre, and Artetxe}]{elhady-etal-2025-emergent}
Ahmed Elhady, Eneko Agirre, and Mikel Artetxe. 2025.
\newblock \href {https://doi.org/10.18653/v1/2025.acl-long.1547} {Emergent abilities of large language models under continued pre-training for language adaptation}.
\newblock In \emph{Proceedings of the 63rd Annual Meeting of the Association for Computational Linguistics (Volume 1: Long Papers)}, pages 32174--32186, Vienna, Austria. Association for Computational Linguistics.

\bibitem[{et~al(2023)}]{workshop2023bloom176bparameteropenaccessmultilingual}
BigScience~Workshop et~al. 2023.
\newblock \href {http://arxiv.org/abs/2211.05100} {Bloom: A 176b-parameter open-access multilingual language model}.

\bibitem[{Etxaniz et~al.(2024{\natexlab{a}})Etxaniz, Azkune, Soroa, de~Lacalle, and Artetxe}]{NEURIPS2024_3bb42f6b}
Julen Etxaniz, Gorka Azkune, Aitor Soroa, Oier~Lopez de~Lacalle, and Mikel Artetxe. 2024{\natexlab{a}}.
\newblock \href {https://proceedings.neurips.cc/paper_files/paper/2024/file/3bb42f6bb1b1ab6809afd6c90865b087-Paper-Datasets_and_Benchmarks_Track.pdf} {Bertaqa: How much do language models know about local culture?}
\newblock In \emph{Advances in Neural Information Processing Systems}, volume~37, pages 34077--34097. Curran Associates, Inc.

\bibitem[{Etxaniz et~al.(2024{\natexlab{b}})Etxaniz, Sainz, Perez, Aldabe, Rigau, Agirre, Ormazabal, Artetxe, and Soroa}]{etxaniz-etal-2024-latxa}
Julen Etxaniz, Oscar Sainz, Naiara Perez, Itziar Aldabe, German Rigau, Eneko Agirre, Aitor Ormazabal, Mikel Artetxe, and Aitor Soroa. 2024{\natexlab{b}}.
\newblock \href {https://doi.org/10.18653/v1/2024.acl-long.799} {Latxa: An open language model and evaluation suite for {B}asque}.
\newblock In \emph{Proceedings of the 62nd Annual Meeting of the Association for Computational Linguistics (Volume 1: Long Papers)}, pages 14952--14972, Bangkok, Thailand. Association for Computational Linguistics.

\bibitem[{Gao et~al.(2024)Gao, Tow, Abbasi, Biderman, Black, DiPofi, Foster, Golding, Hsu, Le~Noac'h, Li, McDonell, Muennighoff, Ociepa, Phang, Reynolds, Schoelkopf, Skowron, Sutawika, Tang, Thite, Wang, Wang, and Zou}]{eval-harness}
Leo Gao, Jonathan Tow, Baber Abbasi, Stella Biderman, Sid Black, Anthony DiPofi, Charles Foster, Laurence Golding, Jeffrey Hsu, Alain Le~Noac'h, Haonan Li, Kyle McDonell, Niklas Muennighoff, Chris Ociepa, Jason Phang, Laria Reynolds, Hailey Schoelkopf, Aviya Skowron, Lintang Sutawika, Eric Tang, Anish Thite, Ben Wang, Kevin Wang, and Andy Zou. 2024.
\newblock \href {https://doi.org/10.5281/zenodo.12608602} {The language model evaluation harness}.

\bibitem[{Goddard et~al.(2024)Goddard, Siriwardhana, Ehghaghi, Meyers, Karpukhin, Benedict, McQuade, and Solawetz}]{goddard-etal-2024-arcees}
Charles Goddard, Shamane Siriwardhana, Malikeh Ehghaghi, Luke Meyers, Vladimir Karpukhin, Brian Benedict, Mark McQuade, and Jacob Solawetz. 2024.
\newblock \href {https://doi.org/10.18653/v1/2024.emnlp-industry.36} {Arcee{'}s {M}erge{K}it: A toolkit for merging large language models}.
\newblock In \emph{Proceedings of the 2024 Conference on Empirical Methods in Natural Language Processing: Industry Track}, pages 477--485, Miami, Florida, US. Association for Computational Linguistics.

\bibitem[{Gonz{\'a}lez et~al.(2026)Gonz{\'a}lez, Borrego-Obrador, Herrero, Sarvazyan, Chinea-Rios, Basile, and Franco-Salvador}]{Gonzalez2025IberBenchLEB}
Jos{\'e}~A. Gonz{\'a}lez, Ian Borrego-Obrador, {\'A}lvaro~Romo Herrero, Areg~Mikael Sarvazyan, Mara Chinea-Rios, Angelo Basile, and Marc Franco-Salvador. 2026.
\newblock \href {https://doi.org/10.1016/j.csl.2025.101899} {Iberbench: Llm evaluation on iberian languages}.
\newblock \emph{Computer Speech \& Language}, 96:101899.

\bibitem[{Gonzalez-Agirre et~al.(2025)Gonzalez-Agirre, P{\`a}mies, Llop, Baucells, Dalt, Tamayo, Saiz, Espu{\~n}a, Prats, Aula-Blasco, Mina, Pikabea, Rubio, Shvets, Sall{\'e}s, Lacunza, Palomar, Falc{\~a}o, Tormo, Vasquez-Reina, Marimon, Pareras, Ruiz-Fern{\'a}ndez, and Villegas}]{gonzalezagirre2025salamandratechnicalreport}
Aitor Gonzalez-Agirre, Marc P{\`a}mies, Joan Llop, Irene Baucells, Severino~Da Dalt, Daniel Tamayo, Jos{\'e}~Javier Saiz, Ferran Espu{\~n}a, Jaume Prats, Javier Aula-Blasco, Mario Mina, I{\~n}igo Pikabea, Adri{\'a}n Rubio, Alexander Shvets, Anna Sall{\'e}s, I{\~n}aki Lacunza, Jorge Palomar, J{\'u}lia Falc{\~a}o, Luc{\'i}a Tormo, Luis Vasquez-Reina, Montserrat Marimon, Oriol Pareras, Valle Ruiz-Fern{\'a}ndez, and Marta Villegas. 2025.
\newblock \href {https://proyectoilenia.es/publicaciones/salamandra-technical-report/} {Salamandra technical report}.
\newblock Technical report, Barcelona Supercomputing Center.

\bibitem[{Goyal et~al.(2022)Goyal, Gao, Chaudhary, Chen, Wenzek, Ju, Krishnan, Ranzato, Guzm{\'a}n, and Fan}]{goyal2021flores101evaluationbenchmarklowresource}
Naman Goyal, Cynthia Gao, Vishrav Chaudhary, Peng-Jen Chen, Guillaume Wenzek, Da~Ju, Sanjana Krishnan, Marc{'}Aurelio Ranzato, Francisco Guzm{\'a}n, and Angela Fan. 2022.
\newblock \href {https://doi.org/10.1162/tacl_a_00474} {The {FLORES}-101 evaluation benchmark for low-resource and multilingual machine translation}.
\newblock \emph{Transactions of the Association for Computational Linguistics}, 10:522--538.

\bibitem[{Grandury et~al.(2025)Grandury, Aula-Blasco, Falc{\~a}o, Fourrier, Saiz, Mart{\'i}nez, Gomez, Agerri, Garc{\'i}a, Chiruzzo, Conde, Gomez~Adorno, Nieto, Ivetta, Fuertes, Plaza-del Arco, Mart{\'i}n-Valdivia, Zamorano, Sanz, Reviriego, Plaza, Vaca~Serrano, Vallecillo-Rodr{\'i}guez, Vallego, and Zubiaga}]{grandury-etal-2025-la}
Mar{\'i}a Grandury, Javier Aula-Blasco, J{\'u}lia Falc{\~a}o, Cl{\'e}mentine Fourrier, Miguel~Gonz{\'a}lez Saiz, Gonzalo Mart{\'i}nez, Gonzalo~Santamaria Gomez, Rodrigo Agerri, Nuria~Aldama Garc{\'i}a, Luis Chiruzzo, Javier Conde, Helena Gomez~Adorno, Marta~Guerrero Nieto, Guido Ivetta, Nat{\`a}lia~L{\'o}pez Fuertes, Flor~Miriam Plaza-del Arco, Mar{\'i}a-Teresa Mart{\'i}n-Valdivia, Helena~Montoro Zamorano, Carmen~Mu{\~n}oz Sanz, Pedro Reviriego, Leire~Rosado Plaza, Alejandro Vaca~Serrano, Estrella Vallecillo-Rodr{\'i}guez, Jorge Vallego, and Irune Zubiaga. 2025.
\newblock \href {https://doi.org/10.18653/v1/2025.acl-long.1561} {La leaderboard: A large language model leaderboard for {S}panish varieties and languages of {S}pain and {L}atin {A}merica}.
\newblock In \emph{Proceedings of the 63rd Annual Meeting of the Association for Computational Linguistics (Volume 1: Long Papers)}, pages 32482--32524, Vienna, Austria. Association for Computational Linguistics.

\bibitem[{Grattafiori et~al.(2024)Grattafiori, Dubey, Jauhri, Pandey, Kadian, Al-Dahle, Letman, Mathur, Schelten, Vaughan, Yang, Fan, Goyal, Hartshorn, Yang, Mitra, Sravankumar, Korenev, Hinsvark, Rao, Zhang, Rodriguez, Gregerson, Spataru, Roziere, Biron, Tang, Chern, Caucheteux, Nayak, Bi, Marra, McConnell, Keller, Touret, Wu, Wong, Ferrer, Nikolaidis, Allonsius, Song, Pintz, Livshits, Wyatt, Esiobu, Choudhary, Mahajan, Garcia-Olano, Perino, Hupkes, Lakomkin, AlBadawy, Lobanova, Dinan, Smith, Radenovic, Guzmán, Zhang, Synnaeve, Lee, Anderson, Thattai, Nail, Mialon, Pang, Cucurell, Nguyen, Korevaar, Xu, Touvron, Zarov, Ibarra, Kloumann, Misra, Evtimov, Zhang, Copet, Lee, Geffert, Vranes, Park, Mahadeokar, Shah, van~der Linde, Billock, Hong, Lee, Fu, Chi, Huang, Liu, Wang, Yu, Bitton, Spisak, Park, Rocca, Johnstun, Saxe, Jia, Alwala, Prasad, Upasani, Plawiak, Li, Heafield, Stone, El-Arini, Iyer, Malik, Chiu, Bhalla, Lakhotia, Rantala-Yeary, van~der Maaten, Chen, Tan, Jenkins, Martin, Madaan, Malo, Blecher,
  Landzaat, de~Oliveira, Muzzi, Pasupuleti, Singh, Paluri, Kardas, Tsimpoukelli, Oldham, Rita, Pavlova, Kambadur, Lewis, Si, Singh, Hassan, Goyal, Torabi, Bashlykov, Bogoychev, Chatterji, Zhang, Duchenne, Çelebi, Alrassy, Zhang, Li, Vasic, Weng, Bhargava, Dubal, Krishnan, Koura, Xu, He, Dong, Srinivasan, Ganapathy, Calderer, Cabral, Stojnic, Raileanu, Maheswari, Girdhar, Patel, Sauvestre, Polidoro, Sumbaly, Taylor, Silva, Hou, Wang, Hosseini, Chennabasappa, Singh, Bell, Kim, Edunov, Nie, Narang, Raparthy, Shen, Wan, Bhosale, Zhang, Vandenhende, Batra, Whitman, Sootla, Collot, Gururangan, Borodinsky, Herman, Fowler, Sheasha, Georgiou, Scialom, Speckbacher, Mihaylov, Xiao, Karn, Goswami, Gupta, Ramanathan, Kerkez, Gonguet, Do, Vogeti, Albiero, Petrovic, Chu, Xiong, Fu, Meers, Martinet, Wang, Wang, Tan, Xia, Xie, Jia, Wang, Goldschlag, Gaur, Babaei, Wen, Song, Zhang, Li, Mao, Coudert, Yan, Chen, Papakipos, Singh, Srivastava, Jain, Kelsey, Shajnfeld, Gangidi, Victoria, Goldstand, Menon, Sharma, Boesenberg,
  Baevski, Feinstein, Kallet, Sangani, Teo, Yunus, Lupu, Alvarado, Caples, Gu, Ho, Poulton, Ryan, Ramchandani, Dong, Franco, Goyal, Saraf, Chowdhury, Gabriel, Bharambe, Eisenman, Yazdan, James, Maurer, Leonhardi, Huang, Loyd, Paola, Paranjape, Liu, Wu, Ni, Hancock, Wasti, Spence, Stojkovic, Gamido, Montalvo, Parker, Burton, Mejia, Liu, Wang, Kim, Zhou, Hu, Chu, Cai, Tindal, Feichtenhofer, Gao, Civin, Beaty, Kreymer, Li, Adkins, Xu, Testuggine, David, Parikh, Liskovich, Foss, Wang, Le, Holland, Dowling, Jamil, Montgomery, Presani, Hahn, Wood, Le, Brinkman, Arcaute, Dunbar, Smothers, Sun, Kreuk, Tian, Kokkinos, Ozgenel, Caggioni, Kanayet, Seide, Florez, Schwarz, Badeer, Swee, Halpern, Herman, Sizov, Guangyi, Zhang, Lakshminarayanan, Inan, Shojanazeri, Zou, Wang, Zha, Habeeb, Rudolph, Suk, Aspegren, Goldman, Zhan, Damlaj, Molybog, Tufanov, Leontiadis, Veliche, Gat, Weissman, Geboski, Kohli, Lam, Asher, Gaya, Marcus, Tang, Chan, Zhen, Reizenstein, Teboul, Zhong, Jin, Yang, Cummings, Carvill, Shepard, McPhie,
  Torres, Ginsburg, Wang, Wu, U, Saxena, Khandelwal, Zand, Matosich, Veeraraghavan, Michelena, Li, Jagadeesh, Huang, Chawla, Huang, Chen, Garg, A, Silva, Bell, Zhang, Guo, Yu, Moshkovich, Wehrstedt, Khabsa, Avalani, Bhatt, Mankus, Hasson, Lennie, Reso, Groshev, Naumov, Lathi, Keneally, Liu, Seltzer, Valko, Restrepo, Patel, Vyatskov, Samvelyan, Clark, Macey, Wang, Hermoso, Metanat, Rastegari, Bansal, Santhanam, Parks, White, Bawa, Singhal, Egebo, Usunier, Mehta, Laptev, Dong, Cheng, Chernoguz, Hart, Salpekar, Kalinli, Kent, Parekh, Saab, Balaji, Rittner, Bontrager, Roux, Dollar, Zvyagina, Ratanchandani, Yuvraj, Liang, Alao, Rodriguez, Ayub, Murthy, Nayani, Mitra, Parthasarathy, Li, Hogan, Battey, Wang, Howes, Rinott, Mehta, Siby, Bondu, Datta, Chugh, Hunt, Dhillon, Sidorov, Pan, Mahajan, Verma, Yamamoto, Ramaswamy, Lindsay, Lindsay, Feng, Lin, Zha, Patil, Shankar, Zhang, Zhang, Wang, Agarwal, Sajuyigbe, Chintala, Max, Chen, Kehoe, Satterfield, Govindaprasad, Gupta, Deng, Cho, Virk, Subramanian, Choudhury,
  Goldman, Remez, Glaser, Best, Koehler, Robinson, Li, Zhang, Matthews, Chou, Shaked, Vontimitta, Ajayi, Montanez, Mohan, Kumar, Mangla, Ionescu, Poenaru, Mihailescu, Ivanov, Li, Wang, Jiang, Bouaziz, Constable, Tang, Wu, Wang, Wu, Gao, Kleinman, Chen, Hu, Jia, Qi, Li, Zhang, Zhang, Adi, Nam, Yu, Wang, Zhao, Hao, Qian, Li, He, Rait, DeVito, Rosnbrick, Wen, Yang, Zhao, and Ma}]{grattafiori2024llama3herdmodels}
Aaron Grattafiori, Abhimanyu Dubey, Abhinav Jauhri, Abhinav Pandey, Abhishek Kadian, Ahmad Al-Dahle, Aiesha Letman, Akhil Mathur, Alan Schelten, Alex Vaughan, Amy Yang, Angela Fan, Anirudh Goyal, Anthony Hartshorn, Aobo Yang, Archi Mitra, Archie Sravankumar, Artem Korenev, Arthur Hinsvark, Arun Rao, Aston Zhang, Aurelien Rodriguez, Austen Gregerson, Ava Spataru, Baptiste Roziere, Bethany Biron, Binh Tang, Bobbie Chern, Charlotte Caucheteux, Chaya Nayak, Chloe Bi, Chris Marra, Chris McConnell, Christian Keller, Christophe Touret, Chunyang Wu, Corinne Wong, Cristian~Canton Ferrer, Cyrus Nikolaidis, Damien Allonsius, Daniel Song, Danielle Pintz, Danny Livshits, Danny Wyatt, David Esiobu, Dhruv Choudhary, Dhruv Mahajan, Diego Garcia-Olano, Diego Perino, Dieuwke Hupkes, Egor Lakomkin, Ehab AlBadawy, Elina Lobanova, Emily Dinan, Eric~Michael Smith, Filip Radenovic, Francisco Guzmán, Frank Zhang, Gabriel Synnaeve, Gabrielle Lee, Georgia~Lewis Anderson, Govind Thattai, Graeme Nail, Gregoire Mialon, Guan Pang,
  Guillem Cucurell, Hailey Nguyen, Hannah Korevaar, Hu~Xu, Hugo Touvron, Iliyan Zarov, Imanol~Arrieta Ibarra, Isabel Kloumann, Ishan Misra, Ivan Evtimov, Jack Zhang, Jade Copet, Jaewon Lee, Jan Geffert, Jana Vranes, Jason Park, Jay Mahadeokar, Jeet Shah, Jelmer van~der Linde, Jennifer Billock, Jenny Hong, Jenya Lee, Jeremy Fu, Jianfeng Chi, Jianyu Huang, Jiawen Liu, Jie Wang, Jiecao Yu, Joanna Bitton, Joe Spisak, Jongsoo Park, Joseph Rocca, Joshua Johnstun, Joshua Saxe, Junteng Jia, Kalyan~Vasuden Alwala, Karthik Prasad, Kartikeya Upasani, Kate Plawiak, Ke~Li, Kenneth Heafield, Kevin Stone, Khalid El-Arini, Krithika Iyer, Kshitiz Malik, Kuenley Chiu, Kunal Bhalla, Kushal Lakhotia, Lauren Rantala-Yeary, Laurens van~der Maaten, Lawrence Chen, Liang Tan, Liz Jenkins, Louis Martin, Lovish Madaan, Lubo Malo, Lukas Blecher, Lukas Landzaat, Luke de~Oliveira, Madeline Muzzi, Mahesh Pasupuleti, Mannat Singh, Manohar Paluri, Marcin Kardas, Maria Tsimpoukelli, Mathew Oldham, Mathieu Rita, Maya Pavlova, Melanie Kambadur,
  Mike Lewis, Min Si, Mitesh~Kumar Singh, Mona Hassan, Naman Goyal, Narjes Torabi, Nikolay Bashlykov, Nikolay Bogoychev, Niladri Chatterji, Ning Zhang, Olivier Duchenne, Onur Çelebi, Patrick Alrassy, Pengchuan Zhang, Pengwei Li, Petar Vasic, Peter Weng, Prajjwal Bhargava, Pratik Dubal, Praveen Krishnan, Punit~Singh Koura, Puxin Xu, Qing He, Qingxiao Dong, Ragavan Srinivasan, Raj Ganapathy, Ramon Calderer, Ricardo~Silveira Cabral, Robert Stojnic, Roberta Raileanu, Rohan Maheswari, Rohit Girdhar, Rohit Patel, Romain Sauvestre, Ronnie Polidoro, Roshan Sumbaly, Ross Taylor, Ruan Silva, Rui Hou, Rui Wang, Saghar Hosseini, Sahana Chennabasappa, Sanjay Singh, Sean Bell, Seohyun~Sonia Kim, Sergey Edunov, Shaoliang Nie, Sharan Narang, Sharath Raparthy, Sheng Shen, Shengye Wan, Shruti Bhosale, Shun Zhang, Simon Vandenhende, Soumya Batra, Spencer Whitman, Sten Sootla, Stephane Collot, Suchin Gururangan, Sydney Borodinsky, Tamar Herman, Tara Fowler, Tarek Sheasha, Thomas Georgiou, Thomas Scialom, Tobias Speckbacher,
  Todor Mihaylov, Tong Xiao, Ujjwal Karn, Vedanuj Goswami, Vibhor Gupta, Vignesh Ramanathan, Viktor Kerkez, Vincent Gonguet, Virginie Do, Vish Vogeti, Vítor Albiero, Vladan Petrovic, Weiwei Chu, Wenhan Xiong, Wenyin Fu, Whitney Meers, Xavier Martinet, Xiaodong Wang, Xiaofang Wang, Xiaoqing~Ellen Tan, Xide Xia, Xinfeng Xie, Xuchao Jia, Xuewei Wang, Yaelle Goldschlag, Yashesh Gaur, Yasmine Babaei, Yi~Wen, Yiwen Song, Yuchen Zhang, Yue Li, Yuning Mao, Zacharie~Delpierre Coudert, Zheng Yan, Zhengxing Chen, Zoe Papakipos, Aaditya Singh, Aayushi Srivastava, Abha Jain, Adam Kelsey, Adam Shajnfeld, Adithya Gangidi, Adolfo Victoria, Ahuva Goldstand, Ajay Menon, Ajay Sharma, Alex Boesenberg, Alexei Baevski, Allie Feinstein, Amanda Kallet, Amit Sangani, Amos Teo, Anam Yunus, Andrei Lupu, Andres Alvarado, Andrew Caples, Andrew Gu, Andrew Ho, Andrew Poulton, Andrew Ryan, Ankit Ramchandani, Annie Dong, Annie Franco, Anuj Goyal, Aparajita Saraf, Arkabandhu Chowdhury, Ashley Gabriel, Ashwin Bharambe, Assaf Eisenman, Azadeh
  Yazdan, Beau James, Ben Maurer, Benjamin Leonhardi, Bernie Huang, Beth Loyd, Beto~De Paola, Bhargavi Paranjape, Bing Liu, Bo~Wu, Boyu Ni, Braden Hancock, Bram Wasti, Brandon Spence, Brani Stojkovic, Brian Gamido, Britt Montalvo, Carl Parker, Carly Burton, Catalina Mejia, Ce~Liu, Changhan Wang, Changkyu Kim, Chao Zhou, Chester Hu, Ching-Hsiang Chu, Chris Cai, Chris Tindal, Christoph Feichtenhofer, Cynthia Gao, Damon Civin, Dana Beaty, Daniel Kreymer, Daniel Li, David Adkins, David Xu, Davide Testuggine, Delia David, Devi Parikh, Diana Liskovich, Didem Foss, Dingkang Wang, Duc Le, Dustin Holland, Edward Dowling, Eissa Jamil, Elaine Montgomery, Eleonora Presani, Emily Hahn, Emily Wood, Eric-Tuan Le, Erik Brinkman, Esteban Arcaute, Evan Dunbar, Evan Smothers, Fei Sun, Felix Kreuk, Feng Tian, Filippos Kokkinos, Firat Ozgenel, Francesco Caggioni, Frank Kanayet, Frank Seide, Gabriela~Medina Florez, Gabriella Schwarz, Gada Badeer, Georgia Swee, Gil Halpern, Grant Herman, Grigory Sizov, Guangyi, Zhang, Guna
  Lakshminarayanan, Hakan Inan, Hamid Shojanazeri, Han Zou, Hannah Wang, Hanwen Zha, Haroun Habeeb, Harrison Rudolph, Helen Suk, Henry Aspegren, Hunter Goldman, Hongyuan Zhan, Ibrahim Damlaj, Igor Molybog, Igor Tufanov, Ilias Leontiadis, Irina-Elena Veliche, Itai Gat, Jake Weissman, James Geboski, James Kohli, Janice Lam, Japhet Asher, Jean-Baptiste Gaya, Jeff Marcus, Jeff Tang, Jennifer Chan, Jenny Zhen, Jeremy Reizenstein, Jeremy Teboul, Jessica Zhong, Jian Jin, Jingyi Yang, Joe Cummings, Jon Carvill, Jon Shepard, Jonathan McPhie, Jonathan Torres, Josh Ginsburg, Junjie Wang, Kai Wu, Kam~Hou U, Karan Saxena, Kartikay Khandelwal, Katayoun Zand, Kathy Matosich, Kaushik Veeraraghavan, Kelly Michelena, Keqian Li, Kiran Jagadeesh, Kun Huang, Kunal Chawla, Kyle Huang, Lailin Chen, Lakshya Garg, Lavender A, Leandro Silva, Lee Bell, Lei Zhang, Liangpeng Guo, Licheng Yu, Liron Moshkovich, Luca Wehrstedt, Madian Khabsa, Manav Avalani, Manish Bhatt, Martynas Mankus, Matan Hasson, Matthew Lennie, Matthias Reso, Maxim
  Groshev, Maxim Naumov, Maya Lathi, Meghan Keneally, Miao Liu, Michael~L. Seltzer, Michal Valko, Michelle Restrepo, Mihir Patel, Mik Vyatskov, Mikayel Samvelyan, Mike Clark, Mike Macey, Mike Wang, Miquel~Jubert Hermoso, Mo~Metanat, Mohammad Rastegari, Munish Bansal, Nandhini Santhanam, Natascha Parks, Natasha White, Navyata Bawa, Nayan Singhal, Nick Egebo, Nicolas Usunier, Nikhil Mehta, Nikolay~Pavlovich Laptev, Ning Dong, Norman Cheng, Oleg Chernoguz, Olivia Hart, Omkar Salpekar, Ozlem Kalinli, Parkin Kent, Parth Parekh, Paul Saab, Pavan Balaji, Pedro Rittner, Philip Bontrager, Pierre Roux, Piotr Dollar, Polina Zvyagina, Prashant Ratanchandani, Pritish Yuvraj, Qian Liang, Rachad Alao, Rachel Rodriguez, Rafi Ayub, Raghotham Murthy, Raghu Nayani, Rahul Mitra, Rangaprabhu Parthasarathy, Raymond Li, Rebekkah Hogan, Robin Battey, Rocky Wang, Russ Howes, Ruty Rinott, Sachin Mehta, Sachin Siby, Sai~Jayesh Bondu, Samyak Datta, Sara Chugh, Sara Hunt, Sargun Dhillon, Sasha Sidorov, Satadru Pan, Saurabh Mahajan,
  Saurabh Verma, Seiji Yamamoto, Sharadh Ramaswamy, Shaun Lindsay, Shaun Lindsay, Sheng Feng, Shenghao Lin, Shengxin~Cindy Zha, Shishir Patil, Shiva Shankar, Shuqiang Zhang, Shuqiang Zhang, Sinong Wang, Sneha Agarwal, Soji Sajuyigbe, Soumith Chintala, Stephanie Max, Stephen Chen, Steve Kehoe, Steve Satterfield, Sudarshan Govindaprasad, Sumit Gupta, Summer Deng, Sungmin Cho, Sunny Virk, Suraj Subramanian, Sy~Choudhury, Sydney Goldman, Tal Remez, Tamar Glaser, Tamara Best, Thilo Koehler, Thomas Robinson, Tianhe Li, Tianjun Zhang, Tim Matthews, Timothy Chou, Tzook Shaked, Varun Vontimitta, Victoria Ajayi, Victoria Montanez, Vijai Mohan, Vinay~Satish Kumar, Vishal Mangla, Vlad Ionescu, Vlad Poenaru, Vlad~Tiberiu Mihailescu, Vladimir Ivanov, Wei Li, Wenchen Wang, Wenwen Jiang, Wes Bouaziz, Will Constable, Xiaocheng Tang, Xiaojian Wu, Xiaolan Wang, Xilun Wu, Xinbo Gao, Yaniv Kleinman, Yanjun Chen, Ye~Hu, Ye~Jia, Ye~Qi, Yenda Li, Yilin Zhang, Ying Zhang, Yossi Adi, Youngjin Nam, Yu, Wang, Yu~Zhao, Yuchen Hao, Yundi
  Qian, Yunlu Li, Yuzi He, Zach Rait, Zachary DeVito, Zef Rosnbrick, Zhaoduo Wen, Zhenyu Yang, Zhiwei Zhao, and Zhiyu Ma. 2024.
\newblock \href {https://ai.meta.com/research/publications/the-llama-3-herd-of-models/} {The llama 3 herd of models}.

\bibitem[{Huang et~al.(2024)Huang, Li, Hsu, Chen, Lin, Hsiao, Tsai, and Lee}]{huang2024chatvector}
Shih-Cheng Huang, Pin-Zu Li, Yu-chi Hsu, Kuang-Ming Chen, Yu~Tung Lin, Shih-Kai Hsiao, Richard Tsai, and Hung-yi Lee. 2024.
\newblock \href {https://doi.org/10.18653/v1/2024.acl-long.590} {Chat vector: A simple approach to equip llms with instruction following and model alignment in new languages}.
\newblock In \emph{Proceedings of the 62nd Annual Meeting of the Association for Computational Linguistics (Volume 1: Long Papers)}, pages 10943--10959, Bangkok, Thailand. Association for Computational Linguistics.

\bibitem[{Ilharco et~al.(2023)Ilharco, Ribeiro, Wortsman, Schmidt, Hajishirzi, and Farhadi}]{ilharco2023editingmodelstaskarithmetic}
Gabriel Ilharco, Marco~Tulio Ribeiro, Mitchell Wortsman, Ludwig Schmidt, Hannaneh Hajishirzi, and Ali Farhadi. 2023.
\newblock \href {https://openreview.net/forum?id=6t0Kwf8-jrj} {Editing models with task arithmetic}.
\newblock In \emph{The Eleventh International Conference on Learning Representations}.
\newblock ICLR 2023 poster.

\bibitem[{Jiang et~al.(2024)Jiang, Sablayrolles, Roux, Mensch, Savary, Bamford, Chaplot, de~las Casas, Hanna, Bressand, Lengyel, Bour, Lample, Lavaud, Saulnier, Lachaux, Stock, Subramanian, Yang, Antoniak, Scao, Gervet, Lavril, Wang, Lacroix, and Sayed}]{jiang2024mixtralexperts}
Albert~Q. Jiang, Alexandre Sablayrolles, Antoine Roux, Arthur Mensch, Blanche Savary, Chris Bamford, Devendra~Singh Chaplot, Diego de~las Casas, Emma~Bou Hanna, Florian Bressand, Gianna Lengyel, Guillaume Bour, Guillaume Lample, Lélio~Renard Lavaud, Lucile Saulnier, Marie-Anne Lachaux, Pierre Stock, Sandeep Subramanian, Sophia Yang, Szymon Antoniak, Teven~Le Scao, Théophile Gervet, Thibaut Lavril, Thomas Wang, Timothée Lacroix, and William~El Sayed. 2024.
\newblock \href {http://arxiv.org/abs/2401.04088} {Mixtral of experts}.

\bibitem[{Le~Scao et~al.(2022)Le~Scao, Wang, Hesslow, Bekman, Bari, Biderman, Elsahar, Muennighoff, Phang, Press, Raffel, Sanh, Shen, Sutawika, Tae, Yong, Launay, and Beltagy}]{le-scao-etal-2022-language}
Teven Le~Scao, Thomas Wang, Daniel Hesslow, Stas Bekman, M~Saiful Bari, Stella Biderman, Hady Elsahar, Niklas Muennighoff, Jason Phang, Ofir Press, Colin Raffel, Victor Sanh, Sheng Shen, Lintang Sutawika, Jaesung Tae, Zheng~Xin Yong, Julien Launay, and Iz~Beltagy. 2022.
\newblock \href {https://doi.org/10.18653/v1/2022.findings-emnlp.54} {What language model to train if you have one million {GPU} hours?}
\newblock In \emph{Findings of the Association for Computational Linguistics: EMNLP 2022}, pages 765--782, Abu Dhabi, United Arab Emirates. Association for Computational Linguistics.

\bibitem[{Lin et~al.(2022)Lin, Mihaylov, Artetxe, Wang, Chen, Simig, Ott, Goyal, Bhosale, Du, Pasunuru, Shleifer, Koura, Chaudhary, O{'}Horo, Wang, Zettlemoyer, Kozareva, Diab, Stoyanov, and Li}]{lin-etal-2022-shot}
Xi~Victoria Lin, Todor Mihaylov, Mikel Artetxe, Tianlu Wang, Shuohui Chen, Daniel Simig, Myle Ott, Naman Goyal, Shruti Bhosale, Jingfei Du, Ramakanth Pasunuru, Sam Shleifer, Punit~Singh Koura, Vishrav Chaudhary, Brian O{'}Horo, Jeff Wang, Luke Zettlemoyer, Zornitsa Kozareva, Mona Diab, Veselin Stoyanov, and Xian Li. 2022.
\newblock \href {https://doi.org/10.18653/v1/2022.emnlp-main.616} {Few-shot learning with multilingual generative language models}.
\newblock In \emph{Proceedings of the 2022 Conference on Empirical Methods in Natural Language Processing}, pages 9019--9052, Abu Dhabi, United Arab Emirates. Association for Computational Linguistics.

\bibitem[{Mihaylov et~al.(2018)Mihaylov, Clark, Khot, and Sabharwal}]{mihaylov-etal-2018-suit}
Todor Mihaylov, Peter Clark, Tushar Khot, and Ashish Sabharwal. 2018.
\newblock \href {https://doi.org/10.18653/v1/D18-1260} {Can a suit of armor conduct electricity? a new dataset for open book question answering}.
\newblock In \emph{Proceedings of the 2018 Conference on Empirical Methods in Natural Language Processing}, pages 2381--2391, Brussels, Belgium. Association for Computational Linguistics.

\bibitem[{Moroni et~al.(2025)Moroni, Aula-Blasco, Conia, Baucells, Perez, Su{\'a}rez, Sall{\'e}s, Ostendorff, Falc{\~a}o, Son, Gonzalez-Agirre, Navigli, and Villegas}]{moroni-etal-2025-multi}
Luca Moroni, Javier Aula-Blasco, Simone Conia, Irene Baucells, Naiara Perez, Silvia~Paniagua Su{\'a}rez, Anna Sall{\'e}s, Malte Ostendorff, J{\'u}lia Falc{\~a}o, Guijin Son, Aitor Gonzalez-Agirre, Roberto Navigli, and Marta Villegas. 2025.
\newblock \href {https://doi.org/10.18653/v1/2025.emnlp-main.1731} {Multi-{LM}entry: Can multilingual {LLM}s solve elementary tasks across languages?}
\newblock In \emph{Proceedings of the 2025 Conference on Empirical Methods in Natural Language Processing}, pages 34126--34157, Suzhou, China. Association for Computational Linguistics.

\bibitem[{Nguyen et~al.(2024{\natexlab{a}})Nguyen, Li, Oh, Schmidt, Weston, Zettlemoyer, and Li}]{nguyen2024betteralignmentinstructionbackandforth}
Thao Nguyen, Jeffrey Li, Sewoong Oh, Ludwig Schmidt, Jason~E. Weston, Luke Zettlemoyer, and Xian Li. 2024{\natexlab{a}}.
\newblock \href {https://doi.org/10.18653/v1/2024.findings-emnlp.777} {Better alignment with instruction back-and-forth translation}.
\newblock In \emph{Findings of the Association for Computational Linguistics: EMNLP 2024}, pages 13289--13308, Miami, Florida, USA. Association for Computational Linguistics.

\bibitem[{Nguyen et~al.(2024{\natexlab{b}})Nguyen, Nguyen, Lai, Man, Ngo, Dernoncourt, Rossi, and Nguyen}]{nguyen-etal-2024-culturax}
Thuat Nguyen, Chien~Van Nguyen, Viet~Dac Lai, Hieu Man, Nghia~Trung Ngo, Franck Dernoncourt, Ryan~A. Rossi, and Thien~Huu Nguyen. 2024{\natexlab{b}}.
\newblock \href {https://aclanthology.org/2024.lrec-main.377/} {{C}ultura{X}: A cleaned, enormous, and multilingual dataset for large language models in 167 languages}.
\newblock In \emph{Proceedings of the 2024 Joint International Conference on Computational Linguistics, Language Resources and Evaluation (LREC-COLING 2024)}, pages 4226--4237, Torino, Italia. ELRA and ICCL.

\bibitem[{OpenAI(2023)}]{openai2024gpt4technicalreport}
OpenAI. 2023.
\newblock \href {https://cdn.openai.com/papers/gpt-4.pdf} {Gpt-4 technical report}.
\newblock Technical report, OpenAI.

\bibitem[{Penedo et~al.(2024)Penedo, Kydl{\'\i}{\v{c}}ek, allal, Lozhkov, Mitchell, Raffel, Werra, and Wolf}]{penedo2024the}
Guilherme Penedo, Hynek Kydl{\'\i}{\v{c}}ek, Loubna~Ben allal, Anton Lozhkov, Margaret Mitchell, Colin Raffel, Leandro~Von Werra, and Thomas Wolf. 2024.
\newblock \href {https://openreview.net/forum?id=n6SCkn2QaG} {The fineweb datasets: Decanting the web for the finest text data at scale}.
\newblock In \emph{The Thirty-eight Conference on Neural Information Processing Systems Datasets and Benchmarks Track}.

\bibitem[{Pipatanakul et~al.(2025)Pipatanakul, Taveekitworachai, Manakul, and Tharnpipitchai}]{pipatanakul2025adaptinglanguagespecificllmsreasoning}
Kunat Pipatanakul, Pittawat Taveekitworachai, Potsawee Manakul, and Kasima Tharnpipitchai. 2025.
\newblock \href {http://arxiv.org/abs/2502.09056} {Adapting language-specific llms to a reasoning model in one day via model merging -- an open recipe}.

\bibitem[{Rae et~al.(2021)Rae, Borgeaud, Cai, Millican, Hoffmann, Song, Aslanides, Henderson, Ring, Young et~al.}]{rae2021scaling}
Jack~W Rae, Sebastian Borgeaud, Trevor Cai, Katie Millican, Jordan Hoffmann, Francis Song, John Aslanides, Sarah Henderson, Roman Ring, Susannah Young, et~al. 2021.
\newblock Scaling language models: Methods, analysis \& insights from training gopher.
\newblock \emph{arXiv preprint arXiv:2112.11446}.

\bibitem[{Raffel et~al.(2020)Raffel, Shazeer, Roberts, Lee, Narang, Matena, Zhou, Li, and Liu}]{raffel2020exloring}
Colin Raffel, Noam Shazeer, Adam Roberts, Katherine Lee, Sharan Narang, Michael Matena, Yanqi Zhou, Wei Li, and Peter~J. Liu. 2020.
\newblock \href {http://jmlr.org/papers/v21/20-074.html} {Exploring the limits of transfer learning with a unified text-to-text transformer}.
\newblock \emph{Journal of Machine Learning Research}, 21(140):1--67.

\bibitem[{Sainz et~al.(2025)Sainz, Perez, Etxaniz, Fernandez~de Landa, Aldabe, Garc{\'i}a-Ferrero, Zabala, Azurmendi, Rigau, Agirre, Artetxe, and Soroa}]{sainz2025instructinglargelanguagemodels}
Oscar Sainz, Naiara Perez, Julen Etxaniz, Joseba Fernandez~de Landa, Itziar Aldabe, Iker Garc{\'i}a-Ferrero, Aimar Zabala, Ekhi Azurmendi, German Rigau, Eneko Agirre, Mikel Artetxe, and Aitor Soroa. 2025.
\newblock \href {https://doi.org/10.18653/v1/2025.emnlp-main.1484} {Instructing large language models for low-resource languages: A systematic study for {B}asque}.
\newblock In \emph{Proceedings of the 2025 Conference on Empirical Methods in Natural Language Processing}, pages 29136--29160, Suzhou, China. Association for Computational Linguistics.

\bibitem[{Sarasua et~al.(2025)Sarasua, Corral, and Saralegi}]{sarasua2025diploma}
Ixak Sarasua, Ander Corral, and Xabier Saralegi. 2025.
\newblock \href {https://doi.org/10.18653/v1/2025.findings-emnlp.1355} {{DIPL}om{A}: Efficient adaptation of instructed {LLM}s to low-resource languages via post-training delta merging}.
\newblock In \emph{Findings of the Association for Computational Linguistics: EMNLP 2025}, pages 24898--24912, Suzhou, China. Association for Computational Linguistics.

\bibitem[{Shliazhko et~al.(2024)Shliazhko, Fenogenova, Tikhonova, Kozlova, Mikhailov, and Shavrina}]{shliazhko2023mgptfewshotlearnersmultilingual}
Oleh Shliazhko, Alena Fenogenova, Maria Tikhonova, Anastasia Kozlova, Vladislav Mikhailov, and Tatiana Shavrina. 2024.
\newblock \href {https://doi.org/10.1162/tacl_a_00633} {m{GPT}: Few-shot learners go multilingual}.
\newblock \emph{Transactions of the Association for Computational Linguistics}, 12:58--79.

\bibitem[{Soldaini et~al.(2024)Soldaini, Kinney, Bhagia, Schwenk, Atkinson, Authur, Bogin, Chandu, Dumas, Elazar, Hofmann, Jha, Kumar, Lucy, Lyu, Lambert, Magnusson, Morrison, Muennighoff, Naik, Nam, Peters, Ravichander, Richardson, Shen, Strubell, Subramani, Tafjord, Walsh, Zettlemoyer, Smith, Hajishirzi, Beltagy, Groeneveld, Dodge, and Lo}]{soldaini-etal-2024-dolma}
Luca Soldaini, Rodney Kinney, Akshita Bhagia, Dustin Schwenk, David Atkinson, Russell Authur, Ben Bogin, Khyathi Chandu, Jennifer Dumas, Yanai Elazar, Valentin Hofmann, Ananya Jha, Sachin Kumar, Li~Lucy, Xinxi Lyu, Nathan Lambert, Ian Magnusson, Jacob Morrison, Niklas Muennighoff, Aakanksha Naik, Crystal Nam, Matthew Peters, Abhilasha Ravichander, Kyle Richardson, Zejiang Shen, Emma Strubell, Nishant Subramani, Oyvind Tafjord, Evan Walsh, Luke Zettlemoyer, Noah Smith, Hannaneh Hajishirzi, Iz~Beltagy, Dirk Groeneveld, Jesse Dodge, and Kyle Lo. 2024.
\newblock \href {https://doi.org/10.18653/v1/2024.acl-long.840} {Dolma: an open corpus of three trillion tokens for language model pretraining research}.
\newblock In \emph{Proceedings of the 62nd Annual Meeting of the Association for Computational Linguistics (Volume 1: Long Papers)}, pages 15725--15788, Bangkok, Thailand. Association for Computational Linguistics.

\bibitem[{Tao et~al.(2024)Tao, Zhang, Huang, Ma, Huang, Zhao, and Feng}]{Tao_2024}
Mingxu Tao, Chen Zhang, Quzhe Huang, Tianyao Ma, Songfang Huang, Dongyan Zhao, and Yansong Feng. 2024.
\newblock \href {https://doi.org/10.18653/v1/2024.findings-emnlp.508} {Unlocking the potential of model merging for low-resource languages}.
\newblock In \emph{Findings of the Association for Computational Linguistics: EMNLP 2024}, page 8705–8720. Association for Computational Linguistics.

\bibitem[{Team and Google(2023)}]{geminiteam2025geminifamilyhighlycapable}
Gemini Team and Google. 2023.
\newblock \href {https://storage.googleapis.com/deepmind-media/gemini/gemini_1_report.pdf} {Gemini: A family of highly capable multimodal models}.
\newblock Technical report.

\bibitem[{{\"U}st{\"u}n et~al.(2024){\"U}st{\"u}n, Aryabumi, Yong, Ko, D{'}souza, Onilude, Bhandari, Singh, Ooi, Kayid, Vargus, Blunsom, Longpre, Muennighoff, Fadaee, Kreutzer, and Hooker}]{ustun2024ayamodelinstructionfinetuned}
Ahmet {\"U}st{\"u}n, Viraat Aryabumi, Zheng Yong, Wei-Yin Ko, Daniel D{'}souza, Gbemileke Onilude, Neel Bhandari, Shivalika Singh, Hui-Lee Ooi, Amr Kayid, Freddie Vargus, Phil Blunsom, Shayne Longpre, Niklas Muennighoff, Marzieh Fadaee, Julia Kreutzer, and Sara Hooker. 2024.
\newblock \href {https://doi.org/10.18653/v1/2024.acl-long.845} {Aya model: An instruction finetuned open-access multilingual language model}.
\newblock In \emph{Proceedings of the 62nd Annual Meeting of the Association for Computational Linguistics (Volume 1: Long Papers)}, pages 15894--15939, Bangkok, Thailand. Association for Computational Linguistics.

\bibitem[{Warstadt et~al.(2019)Warstadt, Singh, and Bowman}]{warstadt-etal-2019-neural}
Alex Warstadt, Amanpreet Singh, and Samuel~R. Bowman. 2019.
\newblock \href {https://doi.org/10.1162/tacl_a_00290} {Neural network acceptability judgments}.
\newblock \emph{Transactions of the Association for Computational Linguistics}, 7:625--641.

\bibitem[{Wortsman et~al.(2022)Wortsman, Ilharco, Gadre, Roelofs, Gontijo-Lopes, Morcos, Namkoong, Farhadi, Carmon, Kornblith, and Schmidt}]{wortsman2022modelsoupsaveragingweights}
Mitchell Wortsman, Gabriel Ilharco, Samir~Ya Gadre, Rebecca Roelofs, Raphael Gontijo-Lopes, Ari~S. Morcos, Hongseok Namkoong, Ali Farhadi, Yair Carmon, Simon Kornblith, and Ludwig Schmidt. 2022.
\newblock \href {https://proceedings.mlr.press/v162/wortsman22a.html} {Model soups: Averaging weights of multiple fine-tuned models improves accuracy without increasing inference time}.
\newblock In \emph{Proceedings of the 39th International Conference on Machine Learning}, volume 162 of \emph{Proceedings of Machine Learning Research}, pages 23965--23998. PMLR.

\bibitem[{Wu et~al.(2023)Wu, Koo, Blum, Black, Kao, Scalzo, and Kurtz}]{wu2023comparativestudyopensourcelarge}
Sean Wu, Michael Koo, Lesley Blum, Andy Black, Liyo Kao, Fabien Scalzo, and Ira Kurtz. 2023.
\newblock \href {http://arxiv.org/abs/2308.04709} {A comparative study of open-source large language models, gpt-4 and claude 2: Multiple-choice test taking in nephrology}.

\bibitem[{Yadav et~al.(2023)Yadav, Tam, Choshen, Raffel, and Bansal}]{Yadav2023TIESMergingRIA}
Prateek Yadav, Derek Tam, Leshem Choshen, Colin~A Raffel, and Mohit Bansal. 2023.
\newblock \href {https://proceedings.neurips.cc/paper_files/paper/2023/file/1644c9af28ab7916874f6fd6228a9bcf-Paper-Conference.pdf} {Ties-merging: Resolving interference when merging models}.
\newblock In \emph{Advances in Neural Information Processing Systems}, volume~36, pages 7093--7115. Curran Associates, Inc.

\bibitem[{Yang et~al.(2025)Yang, Li, Yang, Zhang, Hui, Zheng, Yu, Gao, Huang, Lv, Zheng, Liu, Zhou, Huang, Hu, Ge, Wei, Lin, Tang, Yang, Tu, Zhang, Yang, Yang, Zhou, Zhou, Lin, Dang, Bao, Yang, Yu, Deng, Li, Xue, Li, Zhang, Wang, Zhu, Men, Gao, Liu, Luo, Li, Tang, Yin, Ren, Wang, Zhang, Ren, Fan, Su, Zhang, Zhang, Wan, Liu, Wang, Cui, Zhang, Zhou, and Qiu}]{yang2025qwen3}
An~Yang, Anfeng Li, Baosong Yang, Beichen Zhang, Binyuan Hui, Bo~Zheng, Bowen Yu, Chang Gao, Chengen Huang, Chenxu Lv, Chujie Zheng, Dayiheng Liu, Fan Zhou, Fei Huang, Feng Hu, Hao Ge, Haoran Wei, Huan Lin, Jialong Tang, Jian Yang, Jianhong Tu, Jianwei Zhang, Jianxin Yang, Jiaxi Yang, Jing Zhou, Jingren Zhou, Junyang Lin, Kai Dang, Keqin Bao, Kexin Yang, Le~Yu, Lianghao Deng, Mei Li, Mingfeng Xue, Mingze Li, Pei Zhang, Peng Wang, Qin Zhu, Rui Men, Ruize Gao, Shixuan Liu, Shuang Luo, Tianhao Li, Tianyi Tang, Wenbiao Yin, Xingzhang Ren, Xinyu Wang, Xinyu Zhang, Xuancheng Ren, Yang Fan, Yang Su, Yichang Zhang, Yinger Zhang, Yu~Wan, Yuqiong Liu, Zekun Wang, Zeyu Cui, Zhenru Zhang, Zhipeng Zhou, and Zihan Qiu. 2025.
\newblock \href {https://qwenlm.github.io/blog/qwen3/} {Qwen3 technical report}.
\newblock Technical report.

\bibitem[{Yang et~al.(2026{\natexlab{a}})Yang, Shen, Guo, Wang, Cao, Zhang, and Tao}]{10.1145/3787849}
Enneng Yang, Li~Shen, Guibing Guo, Xingwei Wang, Xiaochun Cao, Jie Zhang, and Dacheng Tao. 2026{\natexlab{a}}.
\newblock \href {https://doi.org/10.1145/3787849} {Model merging in llms, mllms, and beyond: Methods, theories, applications, and opportunities}.
\newblock \emph{ACM Comput. Surv.}, 58(8).

\bibitem[{Yang et~al.(2026{\natexlab{b}})Yang, Shen, Guo, Wang, Cao, Zhang, and Tao}]{yang2024modelmergingllmsmllms}
Enneng Yang, Li~Shen, Guibing Guo, Xingwei Wang, Xiaochun Cao, Jie Zhang, and Dacheng Tao. 2026{\natexlab{b}}.
\newblock \href {https://doi.org/10.1145/3787849} {Model merging in llms, mllms, and beyond: Methods, theories, applications, and opportunities}.
\newblock \emph{ACM Computing Surveys}, 58(8).

\bibitem[{Yu et~al.(2024{\natexlab{a}})Yu, Yu, Yu, Huang, and Li}]{Yu2023LanguageMAA}
Le~Yu, Bowen Yu, Haiyang Yu, Fei Huang, and Yongbin Li. 2024{\natexlab{a}}.
\newblock \href {https://dl.acm.org/doi/10.5555/3692070.3694452} {Language models are super mario: Absorbing abilities from homologous models as a free lunch}.
\newblock In \emph{Proceedings of the 41st International Conference on Machine Learning}, ICML'24. JMLR.org.

\bibitem[{Yu et~al.(2024{\natexlab{b}})Yu, Yu, Yu, Huang, and Li}]{10.5555/3692070.3694452}
Le~Yu, Bowen Yu, Haiyang Yu, Fei Huang, and Yongbin Li. 2024{\natexlab{b}}.
\newblock Language models are super mario: absorbing abilities from homologous models as a free lunch.
\newblock In \emph{Proceedings of the 41st International Conference on Machine Learning}, ICML'24. JMLR.org.

\bibitem[{Zhao et~al.(2023)Zhao, Gu, Varma, Luo, Huang, Xu, Wright, Shojanazeri, Ott, Shleifer, Desmaison, Balioglu, Damania, Nguyen, Chauhan, Hao, Mathews, and Li}]{zhao2023pytorchfsdpexperiencesscaling}
Yanli Zhao, Andrew Gu, Rohan Varma, Liang Luo, Chien-Chin Huang, Min Xu, Less Wright, Hamid Shojanazeri, Myle Ott, Sam Shleifer, Alban Desmaison, Can Balioglu, Pritam Damania, Bernard Nguyen, Geeta Chauhan, Yuchen Hao, Ajit Mathews, and Shen Li. 2023.
\newblock \href {https://doi.org/10.14778/3611540.3611569} {Pytorch {FSDP}: Experiences on scaling fully sharded data parallel}.
\newblock \emph{Proceedings of the VLDB Endowment}, 16(12):3848--3860.

\bibitem[{Zhou et~al.(2023)Zhou, Lu, Mishra, Brahma, Basu, Luan, Zhou, and Hou}]{zhou2023instructionfollowingevaluationlargelanguage}
Jeffrey Zhou, Tianjian Lu, Swaroop Mishra, Siddhartha Brahma, Sujoy Basu, Yi~Luan, Denny Zhou, and Le~Hou. 2023.
\newblock \href {http://arxiv.org/abs/2311.07911} {Instruction-following evaluation for large language models}.

\end{thebibliography}


\appendix
\section{Additional results}

\begin{table*}[!ht]
    \centering
    \resizebox{\textwidth}{!}{
    \begin{tabular}{l|ccccc|ccccc}
        \toprule 
         & \multicolumn{5}{c|}{\textbf{Benchmark average}} & \multicolumn{5}{c}{\textbf{Machine Translation}} \\
        \textbf{Model} & \textbf{\texttt{EU}} & \textbf{\texttt{GL}} & \textbf{\texttt{CA}} & \textbf{\texttt{ES}} & \textbf{\texttt{EN}} & \textbf{\texttt{*-EU}} & \textbf{\texttt{*-GL}} & \textbf{\texttt{*-CA}} & \textbf{\texttt{*-ES}} & \textbf{\texttt{*-EN}} \\ 
        \midrule
        
         Llama 3.1 8B\textsubscript{ joint-EU} & $61.75$ & $58.13$ & $57.81$ & $64.59$ & $73.71$ & $15.03$ & $25.71$ & $29.00$ & $23.86$ & $35.42$ \\
         Salamandra 2B\textsubscript{ Instruct} & $27.95$ & $37.11$ & $43.18$ & $34.68$ & $37.13$ & $6.69$ & $25.27$ & $28.41$ & $21.69$ & $31.50$ \\
         Salamandra 7B\textsubscript{ Instruct} & $44.94$ & $53.60$ & $56.55$ & $52.79$ & $57.27$ & $11.31$ & $\underline{29.94}$ & $\underline{34.12}$ & $25.46$ & $37.67$ \\
         ALIA 40B\textsubscript{ Instruct} & $60.64$ & $\underline{64.98}$ & $\underline{64.68}$ & $62.93$ & $66.04$ & $\underline{15.78}$ & $29.86$ & $33.10$ & $\underline{26.56}$ & $\underline{37.85}$ \\ \midrule
        Llama 3.1 8B\textsubscript{ Instruct} & $49.29$ & $60.11$ & $61.56$ & $\underline{68.22}$ & $73.87$ & $7.18$ & $26.55$ & $30.12$ & $24.01$ & $35.49$ \\
        Llama 3.1 8B\textsubscript{ merge-EU} & $\mathbf{61.48}$ & $57.93$ & $57.40$ & $66.42$ & $\underline{\mathbf{74.83}}$ & $\mathbf{13.62}$ & $25.06$ & $28.14$ & $23.61$ & $34.33$ \\
        Llama 3.1 8B\textsubscript{ merge-GL} & $45.11$ & $\mathbf{61.67}$ & $58.42$ & $67.05$ & $71.86$ & $5.08$ & $\mathbf{28.26}$ & $25.29$ & $21.61$ & $33.19$ \\
        Llama 3.1 8B\textsubscript{ merge-CA} & $45.84$ & $57.71$ & $\mathbf{63.62}$ & $67.03$ & $72.39$ & $5.64$ & $22.95$ & $\mathbf{32.76}$ & $22.93$ & $32.17$ \\
        Llama 3.1 8B\textsubscript{ merge-ES} & $46.71$ & $59.36$ & $59.91$ & $67.61$ & $72.84$ & $6.52$ & $25.47$ & $28.93$ & $23.78$ & $34.45$ \\
        Llama 3.1 8B\textsubscript{ merge-multi} & $53.31$ & $60.19$ & $61.61$ & $\mathbf{67.93}$ & $73.59$ & $8.38$ & $26.00$ & $28.72$ & $\mathbf{24.07}$ & $\mathbf{34.96}$ \\ \midrule
        
        Qwen3 8B\textsubscript{ Instruct} & $44.06$ & $56.51$ & $59.37$ & $64.04$ & $69.84$ & $3.51$ & $24.39$ & $27.78$ & $23.05$ & $33.09$ \\
        Qwen3 8B\textsubscript{ merge-EU} & $\mathbf{55.71}$ & $52.57$ & $56.29$ & $\mathbf{64.10}$ & $\mathbf{72.15}$ & $\mathbf{12.65}$ & $21.84$ & $25.64$ & $22.70$ & $\mathbf{34.09}$ \\
        Qwen3 8B\textsubscript{ merge-GL} & $42.05$ & $\mathbf{57.26}$ & $55.62$ & $61.71$ & $69.58$ & $2.66$ & $\mathbf{29.27}$ & $23.00$ & $21.66$ & $32.85$ \\
        Qwen3 8B\textsubscript{ merge-CA} & $39.73$ & $48.97$ & $58.14$ & $58.90$ & $69.53$ & $2.74$ & $19.79$ & $\mathbf{32.23}$ & $22.05$ & $32.95$ \\
        Qwen3 8B\textsubscript{ merge-ES} & $39.61$ & $50.96$ & $54.39$ & $63.55$ & $70.65$ & $3.17$ & $24.41$ & $26.95$ & $23.36$ & $33.07$ \\
        Qwen3 8B\textsubscript{ merge-multi} & $45.90$ & $54.85$ & $\mathbf{58.67}$ & $62.91$ & $70.26$ & $5.69$ & $26.22$ & $28.89$ & $\mathbf{23.43}$ & $33.89$ \\ \midrule
        
        Qwen3 14B\textsubscript{ Instruct} & $52.09$ & $62.39$ & $62.14$ & $67.87$ & $71.82$ & $5.40$ & $25.87$ & $29.11$ & $24.20$ & $35.10$ \\
        Qwen3 14B\textsubscript{ merge-EU} & $\underline{\mathbf{65.19}}$ & $59.20$ & $60.11$ & $65.58$ & $\mathbf{74.40}$ & $\mathbf{13.90}$ & $24.31$ & $27.61$ & $\mathbf{24.49}$ & $35.81$ \\
        Qwen3 14B\textsubscript{ merge-GL} & $50.53$ & $\mathbf{63.92}$ & $59.04$ & $65.50$ & $71.78$ & $4.60$ & $\mathbf{29.30}$ & $26.48$ & $23.33$ & $34.67$ \\
        Qwen3 14B\textsubscript{ merge-CA} & $51.15$ & $58.62$ & $\mathbf{64.25}$ & $64.33$ & $71.45$ & $4.31$ & $22.07$ & $\mathbf{32.10}$ & $23.66$ & $35.03$ \\
        Qwen3 14B\textsubscript{ merge-ES} & $51.29$ & $61.46$ & $62.11$ & $\mathbf{67.26}$ & $72.67$ & $5.14$ & $25.88$ & $28.90$ & $24.39$ & $35.11$ \\
        Qwen3 14B\textsubscript{ merge-multi} & $56.15$ & $61.80$ & $62.08$ & $66.35$ & $72.72$ & $7.90$ & $27.16$ & $30.32$ & $24.45$ & $\mathbf{35.82}$ \\
        \bottomrule
    \end{tabular}
    }
    \caption{Additional results for Task Arithmetic merges on multiple-choice benchmarks (Accuracy) and machine translation (BLEU). We report baseline models together with Task Arithmetic merge variants for Llama 3.1 8B, Qwen3 8B, and Qwen3 14B. Bold indicates the best result among merged variants of the same backbone model, and underline indicates the best overall result.}
    \label{tab:tabla-ta}
    \vspace{-.4em}
\end{table*}

\begin{table*}[!ht]
    \centering
    \resizebox{\textwidth}{!}{
    \begin{tabular}{l|ccccc|ccccc}
        \toprule
          & \multicolumn{5}{c|}{\textbf{Benchmark average}} & \multicolumn{5}{c}{\textbf{Machine Translation}} \\
         \textbf{Model} & \textbf{\texttt{EU}} & \textbf{\texttt{GL}} & \textbf{\texttt{CA}} & \textbf{\texttt{ES}} & \textbf{\texttt{EN}} & \textbf{\texttt{*-EU}} & \textbf{\texttt{*-GL}} & \textbf{\texttt{*-CA}} & \textbf{\texttt{*-ES}} & \textbf{\texttt{*-EN}} \\ \midrule

         Llama 3.1 8B\textsubscript{ joint-EU} & $61.75$ & $58.13$ & $57.81$ & $64.59$ & $73.71$ & $15.03$ & $25.71$ & $29.00$ & $23.86$ & $35.42$ \\
         Salamandra 2B\textsubscript{ Instruct} & $27.95$ & $37.11$ & $43.18$ & $34.68$ & $37.13$ & $6.69$ & $25.27$ & $28.41$ & $21.69$ & $31.50$ \\
         Salamandra 7B\textsubscript{ Instruct} & $44.94$ & $53.60$ & $56.55$ & $52.79$ & $57.27$ & $11.31$ & $\underline{29.94}$ & $\underline{34.12}$ & $25.46$ & $37.67$ \\
         ALIA 40B\textsubscript{ Instruct} & $60.64$ & $\underline{64.98}$ & $64.68$ & $62.93$ & $66.04$ & $\underline{15.78}$ & $29.86$ & $33.10$ & $\underline{26.56}$ & $\underline{37.85}$ \\ \midrule
         
         Llama 3.1 8B\textsubscript{ Instruct} & $49.29$ & $60.11$ & $61.56$ & $68.22$ & $73.87$ & $7.18$ & $26.55$ & $30.12$ & $24.01$ & $35.49$ \\
         Llama 3.1 8B\textsubscript{ merge-EU nearswap} & $\mathbf{60.43}$ & $56.19$ & $56.77$ & $64.37$ & $\underline{\mathbf{74.85}}$ & $\mathbf{14.56}$ & $24.81$ & $28.60$ & $23.59$ & $36.61$ \\
         Llama 3.1 8B\textsubscript{ merge-GL linear} & $47.56$ & $\mathbf{63.94}$ & $59.96$ & $68.23$ & $73.56$ & $6.37$ & $\mathbf{28.91}$ & $27.81$ & $23.76$ & $36.69$ \\
         Llama 3.1 8B\textsubscript{ merge-CA linear} & $48.26$ & $59.84$ & $\mathbf{63.99}$ & $67.87$ & $73.48$ & $6.85$ & $24.49$ & $\mathbf{32.92}$ & $23.79$ & $36.77$ \\
         Llama 3.1 8B\textsubscript{ merge-ES linear} & $40.46$ & $60.26$ & $60.62$ & $\mathbf{68.87}$ & $74.24$ & $7.65$ & $26.91$ & $30.84$ & $\mathbf{24.46}$ & $36.71$ \\
         Llama 3.1 8B\textsubscript{ merge-multi linear} & $51.66$ & $62.23$ & $61.72$ & $68.46$ & $74.06$ & $8.42$ & $27.19$ & $31.10$ & $24.31$ & $\mathbf{37.02}$ \\ \midrule

         Qwen3 8B\textsubscript{ Instruct} & $44.06$ & $56.51$ & $59.37$ & $64.04$ & $69.84$ & $3.51$ & $24.39$ & $27.78$ & $23.05$ & $33.09$ \\
         Qwen3 8B\textsubscript{ merge-EU ta} & $\mathbf{55.71}$ & $52.57$ & $56.29$ & $64.10$ & $72.15$ & $\mathbf{12.65}$ & $21.84$ & $25.64$ & $22.70$ & $\mathbf{34.09}$ \\
         Qwen3 8B\textsubscript{ merge-GL ta} & $42.05$ & $\mathbf{57.26}$ & $55.62$ & $61.71$ & $69.58$ & $2.66$ & $\mathbf{29.27}$ & $23.00$ & $21.66$ & $32.85$ \\
         Qwen3 8B\textsubscript{ merge-CA ta} & $39.73$ & $48.97$ & $58.14$ & $58.90$ & $69.53$ & $2.74$ & $19.79$ & $\mathbf{32.23}$ & $22.05$ & $32.95$ \\
         Qwen3 8B\textsubscript{ merge-ES linear} & $41.03$ & $56.74$ & $\mathbf{60.26}$ & $\mathbf{65.91}$ & $\mathbf{72.44}$ & $3.49$ & $25.00$ & $28.00$ & $\mathbf{23.90}$ & $33.32$ \\
         Qwen3 8B\textsubscript{ merge-multi ta} & $45.90$ & $54.85$ & $58.67$ & $62.91$ & $70.26$ & $5.69$ & $26.22$ & $28.89$ & $23.43$ & $33.89$ \\ \midrule

         Qwen3 14B\textsubscript{ Instruct} & $52.09$ & $62.39$ & $62.14$ & $67.87$ & $71.82$ & $5.40$ & $25.87$ & $29.11$ & $24.20$ & $35.10$ \\
         Qwen3 14B\textsubscript{ merge-EU ta} & $\underline{\mathbf{65.19}}$ & $59.20$ & $60.11$ & $65.58$ & $74.40$ & $\mathbf{13.90}$ & $24.31$ & $27.61$ & $24.49$ & $35.81$ \\
         Qwen3 14B\textsubscript{ merge-GL ta} & $50.53$ & $63.92$ & $59.04$ & $65.50$ & $71.78$ & $4.60$ & $\mathbf{29.30}$ & $26.48$ & $23.33$ & $34.67$ \\
         Qwen3 14B\textsubscript{ merge-CA linear} & $53.85$ & $63.28$ & $\underline{\mathbf{66.84}}$ & $69.39$ & $74.50$ & $5.49$ & $25.92$ & $\mathbf{32.40}$ & $24.54$ & $36.22$ \\
         Qwen3 14B\textsubscript{ merge-ES linear} & $53.52$ & $64.02$ & $64.90$ & $\underline{\mathbf{69.98}}$ & $\mathbf{74.78}$ & $5.86$ & $27.04$ & $30.02$ & $24.78$ & $36.21$ \\
         Qwen3 14B\textsubscript{ merge-multi linear} & $56.37$ & $\mathbf{64.15}$ & $65.17$ & $69.47$ & $74.77$ & $7.14$ & $27.63$ & $30.91$ & $\mathbf{24.88}$ & $\mathbf{36.45}$ \\
         \bottomrule
    \end{tabular}
    }
    \caption{Additional results showing the best-performing merge configuration per language and backbone on multiple-choice benchmarks (Accuracy) and machine translation (BLEU). For each merged model, we report the strongest variant among the explored merge methods. Bold indicates the best result among merged variants of the same backbone model, and underline indicates the best overall result.}
    \label{tab:main-results}
    \vspace{-.4em}
\end{table*}

\end{document}